\documentclass[11pt,a4paper]{article}

\usepackage[utf8]{inputenc}
\usepackage[T1]{fontenc}
\usepackage[margin=2.5cm]{geometry}
\usepackage{graphicx}
\usepackage{hyperref}
\usepackage{authblk}
\usepackage{xcolor}
\usepackage{multirow}
\usepackage{longtable}
\usepackage{booktabs}
\usepackage{amsmath, amssymb}
\usepackage{bm}
\usepackage{mathtools}
\usepackage{siunitx}
\usepackage{subcaption}
\usepackage{enumitem}
\usepackage{float}

\usepackage[
  backend=biber,
  style=vancouver,
  citestyle=numeric-comp,
  sortcites=true,  %
]{biblatex}

\graphicspath{{fig/}}

\hypersetup{
    colorlinks=true,
    linkcolor=blue,
    citecolor=blue,
    urlcolor=cyan,
}

\newcommand{\dd}{\mathop{}\!\mathrm{d}}

\title{\textbf{A Deep Reinforcement Learning Framework for Closed-loop Guidance of Fish Schools\\via Virtual Agents}}

\author{Takato Shibayama}
\author{Hiroaki Kawashima\thanks{Corresponding author: kawashima@gsis.u-hyogo.ac.jp}}

\affil{Graduate School of Information Science, University of Hyogo, Kobe, Japan}

\date{}

\begin{document}

\maketitle

\begin{abstract}
Guiding collective motion in biological groups is a fundamental challenge in understanding social interaction rules.
In this study, we propose a deep reinforcement learning (RL) framework for closed-loop guidance of fish schools using virtual agents.
These agents are controlled by policies trained via Proximal Policy Optimization (PPO) in simulation and deployed in physical experiments with rummy-nose tetras (\textit{Petitella bleheri}), enabling real-time interaction between artificial agents and live individuals.
To cope with the stochastic behavior of live individuals, we designed a composite reward function that balances directional guidance with cohesion, providing a form of functional biomimicry at the level of the control objective.
Our systematic evaluation of visual parameters showed that a white background and larger stimulus sizes produced the highest guidance efficacy among the tested conditions in physical trials.
Furthermore, evaluation across group sizes and agent configurations indicated that guidance efficacy decreased as the group size increased from five to eight individuals, and that using multiple independently controlled agents did not improve guidance.
Analysis of agent motion in the physical trials indicated that, under the learned policy, the agent moved toward the target while remaining close to the school and re-approached the school after advancing too far ahead.
This study highlights the potential of deep RL for closed-loop guidance of fish schools and identifies challenges in maintaining artificial influence in larger groups.
\end{abstract}

\section{Introduction}
\label{sec:introduction}

Collective behavior in biological systems, such as fish schooling and bird flocking, emerges from local interactions among individuals and gives rise to coordinated group-level patterns~\cite{COUZIN2002,Julia_K,balleriniInteractionRulingAnimal2008}.
Such local interactions enable groups to coordinate their movements and propagate changes in motion among group members~\cite{herbert-readInferringRulesInteraction2011,katzInferringStructureDynamics2011a}. They can also lead to collective decision-making and transitions between distinct group-level states~\cite{COUZIN2002,Couzin2005,tunstromCollectiveStatesMultistability2013}.
Understanding these interaction mechanisms is a central problem in collective behavior research and can also inform the development of fish-inspired robotic systems~\cite{aureliPortraitsSelforganizationFish2012,shintakeBiomimeticUnderwaterRobots2016,katzschmannExplorationUnderwaterLife2018,aritaniSmallRoboticFish2019,berlingerImplicitCoordination3D2021}.

To investigate collective behavior using controlled external stimuli, researchers have employed biomimetic robots~\cite{swain,aureliPortraitsSelforganizationFish2012,marrasFishRobotsSwimming2012,butailCollectiveResponseZebrafish2013,landgrafRoboFishIncreasedAcceptance2016,bonnetInfiltratingZebrafishSwarm2016,liVortexPhaseMatching2020,bierbachLiveFishLearn2022,papaspyrosQuantifyingBiomimicryGap2024b}, as well as visual conspecific representations displayed on screens or through projection~\cite{nakayasuBiologicalMotionStimuli2014,Kawashima14,lemassonMotionCuesTune2018,larschBiologicalMotionInnate2018,liReverseEngineeringControl2025}.
Because their appearance and motion can be systematically controlled, these artificial agents provide a means of isolating specific social cues and examining their causal effects on live animals~\cite{marrasFishRobotsSwimming2012,landgrafRoboFishIncreasedAcceptance2016,larschBiologicalMotionInnate2018,cazenilleHowBlendRobot2018,lemassonMotionCuesTune2018,liReverseEngineeringControl2025}.
In particular, closed-loop systems, in which an artificial agent responds in real time to the behavior of the animals, enable reciprocal interaction between artificial agents and biological individuals~\cite{swain,Kopman,landgrafBlendingShoalRobotic2014a,landgrafRoboFishIncreasedAcceptance2016,bonnetClosedloopInteractionsShoal2018,cazenilleHowBlendRobot2018,chemtobStrategiesModulateZebrafish2020,maxeinerSocialCompetenceImproves2023,papaspyrosQuantifyingBiomimicryGap2024b,liReverseEngineeringControl2025}.
Such systems offer a promising basis not only for studying social interaction mechanisms but also for modulating or actively guiding collective motion~\cite{landgrafBlendingShoalRobotic2014a,bonnetClosedloopInteractionsShoal2018,chemtobStrategiesModulateZebrafish2020,maxeinerSocialCompetenceImproves2023}.

Designing an effective closed-loop controller for a biological collective remains challenging because animal responses are stochastic, nonlinear, and dependent on the evolving social configuration of the group.
Machine-learning approaches have been used to infer social interaction dynamics from fish-pair trajectories and to deploy the resulting models on robotic lures to reproduce and evaluate biomimetic fish--robot interactions~\cite{papaspyrosPredictingLongtermCollective2024,papaspyrosQuantifyingBiomimicryGap2024b}.
Complementing these approaches, reinforcement learning (RL) enables an artificial agent to optimize an action policy with respect to an explicit task objective.
RL has been applied to the real-time control of an artificial visual stimulus interacting with an individual medaka~\cite{nishimuraEstablishingInteractionMachine2016}, the simulation of adaptive swimming behavior in virtual fish~\cite{liSimulatingFishAutonomous2025}, and robotic leadership policies used for embodied model evaluation in dyadic fish--robot interactions~\cite{hockeRobotsThatLearn2026}.
Beyond applications to autonomous swimming and dyadic animal--machine interaction, RL-based guidance has also been demonstrated for a three-fish school using Q-learning-controlled virtual agents~\cite{Nishii18-26}.

More broadly, studies using robotic and virtual conspecifics have shown that social influence depends on factors such as spatial proximity, movement coordination, responsiveness, and integration into a group~\cite{landgrafBlendingShoalRobotic2014a,cazenilleHowMimeticShould2018,chemtobStrategiesModulateZebrafish2020,maxeinerSocialCompetenceImproves2023}. These findings suggest that an artificial agent must remain socially coupled with a biological collective to exert sustained influence on the collective's motion.
From this functional perspective, we developed a deep RL framework for closed-loop guidance of fish schools via virtual agents.
Specifically, we designed a composite reward that balances cohesion with directional guidance by introducing a form of functional biomimicry at the level of the control objective.
The framework also employs Proximal Policy Optimization (PPO)~\cite{ppo} with a continuous state representation, providing greater flexibility than the discretized state formulation used in previous work~\cite{Nishii18-26}.
We hypothesized that optimizing the policy using both reward terms would enable the agent to remain within an effective interaction range while inducing target-directed collective motion, thereby producing more reliable guidance under stochastic fish responses.
The policy was trained entirely in a stochastic simulation environment, evaluated against the previously used single-reward formulation, and deployed without further learning in real-time closed-loop experiments with live fish.

We evaluated the framework with rummy-nose tetras (\textit{Petitella bleheri}) in two stages. First, using schools of $N_r=3$, with $N_r$ denoting the number of real fish, we systematically examined how visual parameters, specifically background color and fish-image size, affected the guidance efficacy of the virtual agents. Second, using schools of $N_r=5$ and $N_r=8$, we evaluated how guidance performance depended on group size and whether independently controlled agents assigned to local subgroups offered an advantage over a fixed-formation agent. 
The results identified visual conditions that support effective guidance and demonstrated significant directional guidance across all agent configurations tested under these conditions. However, guidance efficacy clearly decreased as the group size increased from five to eight individuals.
This work demonstrates the potential of simulation-trained deep RL for biomimetic closed-loop interactions with living collectives, while revealing that maintaining artificial influence becomes more difficult as the group size increases.

\section{Methods}

\subsection{Experimental setup}
We used rummy-nose tetras obtained from a commercial supplier as \textit{Hemigrammus bleheri} (recently reclassified as \textit{Petitella bleheri}). This species was selected due to its strong schooling tendency and its suitability for laboratory maintenance. The fish used in the experiments were approximately \qty{3.5}{\cm} in length.

\begin{figure}[t]
  \centering
  \includegraphics[width=0.9\linewidth]{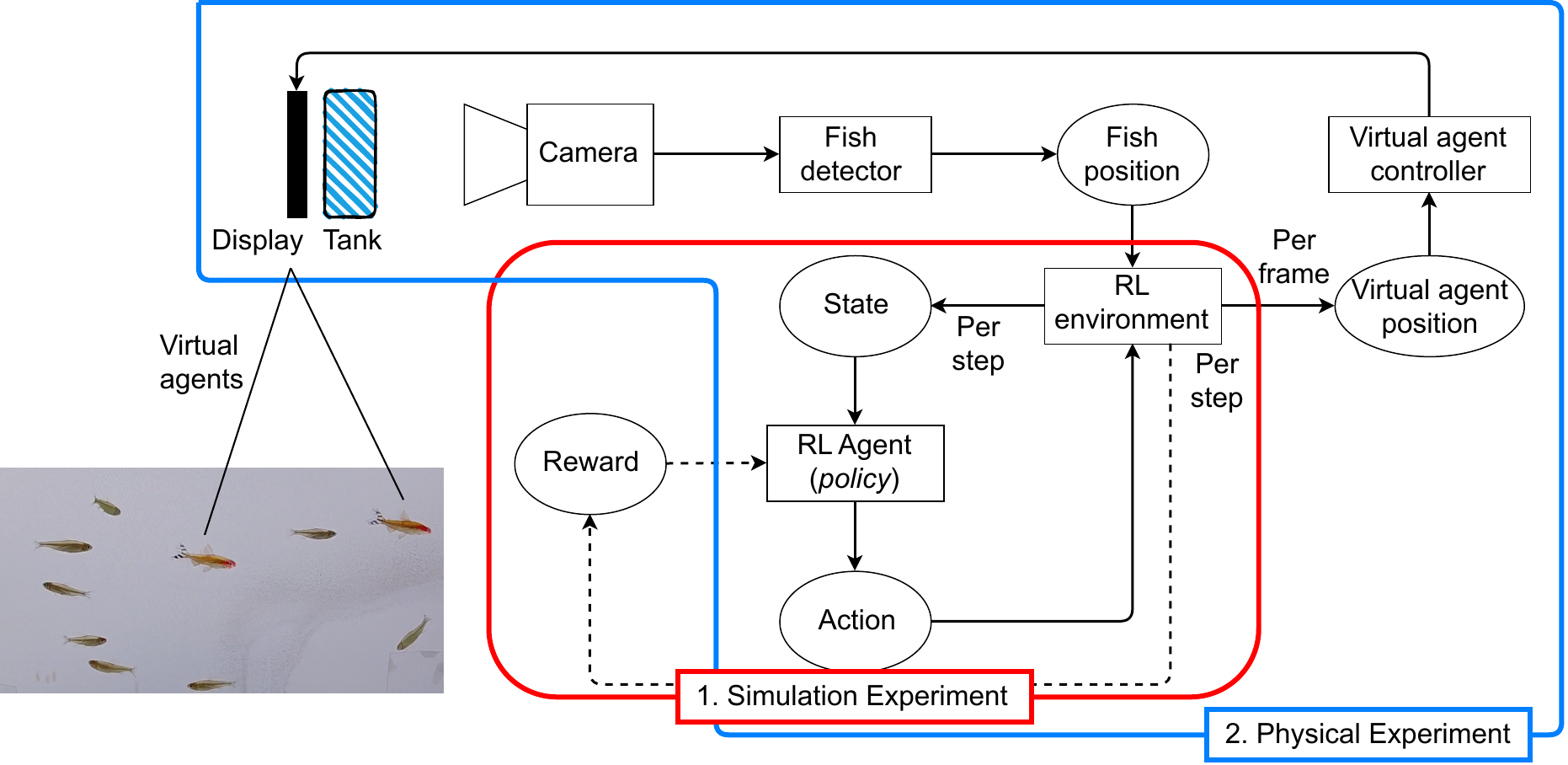}
  \caption{Schematic diagram of the closed-loop system architecture. The positions of the live fish are monitored by a front-facing camera and processed by a PC to apply the learned agent policies, which are then rendered as virtual agents on the display.}
  \label{fig:system_architecture}
\end{figure}

The closed-loop guidance system developed in this study integrated real-time visual tracking of live fish with the application of control policies for virtual agents, as illustrated in the system architecture (Fig.~\ref{fig:system_architecture}). A front-facing camera (Camera: ELP-USBFHD08S-MFV, 1280$\times$720, 120~fps; Lens: Kowa LM3NCM-WP, 3.5mm) captured the positions of the live fish, which were processed by a desktop PC (Intel Core i9-10900K, 64GB RAM, NVIDIA RTX 3080 GPU) to determine the movements of the virtual agents based on trained reinforcement learning policies. These agents were then presented to the fish in real-time via a liquid crystal display.
The camera and display were aligned parallel to the tank surface, minimizing geometric and perspective distortions between the image and display coordinate systems.

\begin{figure}[t]
  \centering
  \includegraphics[width=0.8\linewidth]{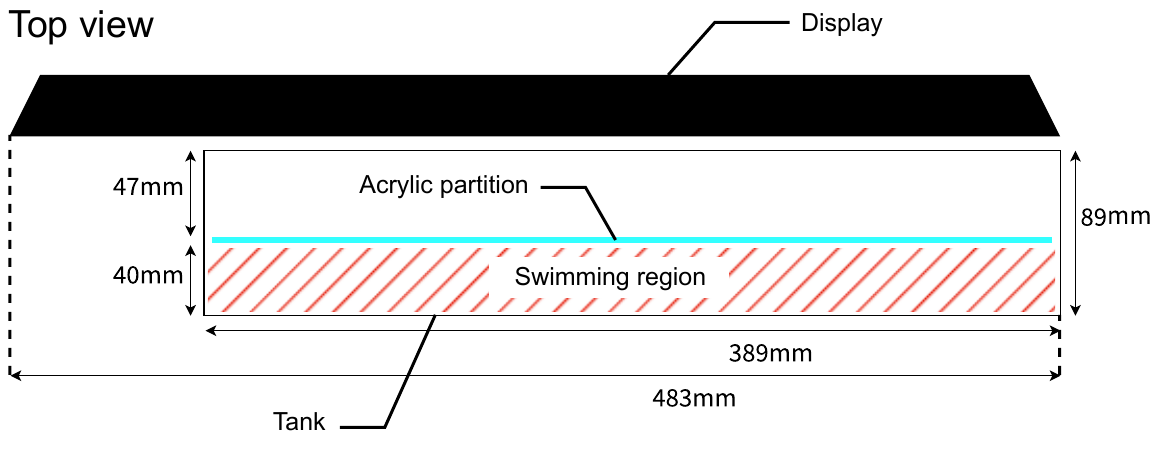}
  \caption{Top-view schematic of the experimental setup. The hatched area indicates the \qty{40}{\mm} deep (front-to-back distance) section where the live fish were constrained. The partition ensures two-dimensional movement and visibility of the displayed virtual agents by avoiding reflections at the tank-water interface from the fish's perspective.}
  \label{fig:tank_setup}
\end{figure}

The experimental arena consisted of an acrylic tank with internal dimensions of \qtyproduct{389 x 213 x 89}{\mm} (width~$\times$~height~$\times$~depth), where depth refers to the front-to-back dimension. As illustrated in the top-view schematic (Fig.~\ref{fig:tank_setup}), we used a \qty{2}{\mm}-thick acrylic partition to divide the tank into two sections with depths of \qty{47}{\mm} and \qty{40}{\mm}; the fish were placed in the \qty{40}{\mm} deep section to constrain their swimming movements to a quasi-two-dimensional plane.
Virtual agents were presented on a liquid crystal display (Philips 221E9/11, \qtyproduct{483 x 270}{\mm}) mounted flush against the rear exterior wall of the tank. This spatial configuration ensured that the swimming region for the live fish was positioned at a sufficient distance from the display to prevent the stimuli from being obscured from the fish's perspective by reflections at the tank-water interface.
Because the display emitted linearly polarized light, a linear polarizing filter (CCS PL-40-NL) with its axis crossed relative to the display was placed on the camera lens, optically removing the virtual agents from the captured images so that only the live fish were tracked.

\subsection{Real-time detection and coordinate mapping}
\label{sec:vision_mapping}

To achieve real-time closed-loop interaction, we implemented an automated detection system using YOLOv8~\cite{yolov8_ultralytics}. The detection model was fine-tuned specifically to identify rummy-nose tetras within the experimental arena. %
All detected individuals in each frame were used to determine the guidance reference points for real-time control (Section~\ref{sec:state_action}).

The raw coordinates $(u_j, v_j)$ of each fish $j \in \{1, \dots, N_r\}$ detected in the camera frame are mapped to a normalized tank coordinate system $\bm{x}^{(r)}_j \in [0, 1]^2$. Before the experiments, the coordinates of the top-left $\bm{u}_{\mathrm{tank}0} = (u_{\mathrm{tank}0}, v_{\mathrm{tank}0})^\top$ and bottom-right $\bm{u}_{\mathrm{tank}1} = (u_{\mathrm{tank}1}, v_{\mathrm{tank}1})^\top$ corners of the swimming region in the camera image were recorded. The normalized position $\bm{x}^{(r)}_j$ is then calculated as follows:
\begin{equation} \label{eq:normalize_real}
  \bm{x}^{(r)}_j = \left(
    \begin{array}{c}
      (u_j - u_{\mathrm{tank}0}) / (u_{\mathrm{tank}1} - u_{\mathrm{tank}0}) \\
      (v_j - v_{\mathrm{tank}0}) / (v_{\mathrm{tank}1} - v_{\mathrm{tank}0})
    \end{array}
  \right).
\end{equation}

To project the virtual agents managed in the RL environment onto the display at the correct physical locations, we establish a mapping between the camera and display coordinate systems. A set of reference spots $\mathcal{S}$ is projected onto the display, and their corresponding homogeneous coordinates in the camera frame, $\mathcal{F} = \{ (u_1, v_1, 1)^\top, \dots, (u_n, v_n, 1)^\top \}$, are captured. The $2 \times 3$ transformation matrix $A$ is then determined by $A = \mathcal{S} \mathcal{F}^{\top} (\mathcal{F} \mathcal{F}^{\top})^{-1}$.

Using this matrix $A$, the display pixel coordinates corresponding to the corners of the swimming region, $\bm{d}_{\mathrm{tank}k} = (d_{\mathrm{tank}k}, e_{\mathrm{tank}k})^\top$ (for $k \in \{0, 1\}$), are obtained by $\bm{d}_{\mathrm{tank}k} = A \tilde{\bm{u}}_{\mathrm{tank}k}$, where $\tilde{\bm{u}}_{\mathrm{tank}k} = (u_{\mathrm{tank}k}, v_{\mathrm{tank}k}, 1)^\top$ represents the homogeneous coordinates of the corners. Finally, the normalized coordinates of virtual agent $i$, denoted as $\bm{x}^{(v)}_i$, are converted into the display pixel coordinates $(d_i, e_i)$ as follows:
\begin{equation} \label{eq:compute_display_virtual}
  \left(
    \begin{array}{c}
      d_i \\ e_i
    \end{array}
  \right) = \bm{x}^{(v)}_i \odot (\bm{d}_{\mathrm{tank}1} - \bm{d}_{\mathrm{tank}0}) + \bm{d}_{\mathrm{tank}0},
\end{equation}
where $\odot$ denotes the element-wise product. This pipeline ensures that the visual stimuli are presented at precisely defined spatial positions relative to the live individuals, enabling accurate guidance based on social interactions.

\subsection{Reinforcement learning framework}
To develop an autonomous control policy for virtual agents capable of guiding fish schools, we employed Proximal Policy Optimization (PPO), a policy-based deep reinforcement learning algorithm.
In this framework, an agent (i.e., a virtual agent presented on the display) is defined as a single policy-controlled unit that interacts with the environment at discrete time steps $t = 0, 1, \dots$ to maximize rewards. Depending on the experimental configuration (see Section~\ref{sec:experimental_design}), the visual representation of an agent was rendered either as a single fish image or as a fixed formation of multiple fish images, where a fish image refers to an individual visual stimulus displayed on the screen.
Each agent operates based on its own state observation and action output.

\subsubsection{State and action space}
\label{sec:state_action}
The state vector observed by each virtual agent $i \in \{1, \dots, N_v\}$ at time $t$ is defined by the coordinates of the real fish and the agent's position:
\begin{equation} \label{eq:state_t}
  \bm{s}_{i,t} = [ \bm{s}_{i,t}^{(r)\top}, \bm{s}_{i,t}^{(v)\top} ]^\top,
\end{equation}
where $\bm{s}_{i,t}^{(v)}$ is the normalized coordinate $\bm{x}_i^{(v)}$ of the $i$-th virtual agent. While prior work represented the real fish collective using its global centroid~\cite{Nishii18-26}, such an approach may lack the granularity required to manage fragmented subgroups. To ensure a scalable and consistent representation of the target collective, we defined the real fish information $\bm{s}_{i,t}^{(r)}$ as a 2D coordinate representing a specific guidance reference point. 

We implemented two modes for defining these reference points based on the experimental configuration:
\begin{itemize}
    \item \textbf{Global mode}: The guidance reference point $\bm{s}_{i,t}^{(r)}$ is defined as the global centroid of the $N_r$ real individuals. In this study, this mode was applied to single-agent scenarios ($N_v = 1$), in which the school was treated as a single cohesive unit.
    \item \textbf{Cluster-assignment mode}: The real fish were partitioned into $k$ clusters using the $k$-means algorithm. 
    At each control step, the correspondence between the virtual agents and cluster centroids was determined using the Hungarian algorithm to minimize the total pairwise distance.
    This mode was applied to our multi-agent configurations ($N_v > 1$), where each virtual agent $i$ is assigned the centroid of a specific cluster $\bm{c}_i$ as its guidance reference point $\bm{s}_{i,t}^{(r)}$. This mapping ensures that each agent maintains a fixed-length input and focuses on a localized subgroup, which was intended to provide robustness against group fragmentation.
\end{itemize}

Each virtual agent $i$ outputs a discrete action $a \in \{0, 1, \dots, 7\}$, which corresponds to eight movement directions equally spaced at $45^\circ$, where $a = 0$ points toward the current target end and increasing $a$ rotates the direction (e.g., $a = 2$ points downward).
Based on the selected action, a target coordinate $\bm{x}_{i, \mathrm{target}}^{(v)}$ is determined for each virtual agent $i$ by displacing its current coordinate by a fixed step of magnitude $D$ along the chosen direction. If the resulting coordinate falls outside the tank, it is reflected back into the valid range $[0, 1]$ at each wall.
In the physical experiments (Section~\ref{sec:physical_experiments}), each fish image was horizontally flipped in real time to align it with its instantaneous movement direction, ensuring a natural visual appearance.
To ensure biologically plausible trajectories, the actual movement of virtual agent $i$ is modeled as a first-order lag system, as follows:
\begin{equation} \label{eq:virtual_move}
  \frac{\dd \bm{x}^{(v)}_i}{\dd t} = \frac{1}{\tau^{(v)}}(\bm{x}_{i, \mathrm{target}}^{(v)} - \bm{x}^{(v)}_i),
\end{equation}
where $\tau^{(v)} > 0$ denotes the time constant. The values of $D$ and $\tau^{(v)}$ used in this study are listed in Supplementary Table~\ref{tab:supp_dynamics_params}.
This dynamic model allows the virtual agent to provide a simplified approximation of the intermittent burst-and-coast movement and frequent directional shifts characteristic of rummy-nose tetras~\cite{caloviDisentanglingModelingInteractions2018}.

\subsubsection{Multi-objective reward design}
To develop an autonomous policy for active guidance, we define a composite reward function $r_{\beta}$. 
We first consider a baseline reward $r_{\mathrm{base}} \in [-1, 1]$, based solely on $c_x^{(r)}$, the horizontal position of the collective's global centroid~\cite{Nishii18-26}:
\begin{equation} \label{eq:r_base}
  r_{\mathrm{base}} = 1 - 2 | c_x^{(r)} - x_{\mathrm{target\text{-}end}} | 
  = \begin{cases}
      2 c_x^{(r)} - 1 & (\text{rightward}) \\
      1 - 2 c_x^{(r)} & (\text{leftward})
    \end{cases},
\end{equation}
where $x_{\mathrm{target\text{-}end}} \in \{0, 1\}$ represents the target end of the tank (0 for leftward and 1 for rightward guidance).
However, using $r_{\mathrm{base}}$ alone can lead to guidance failure.
Since rummy-nose tetras are naturally exploratory, they might move toward the target area independently. In such cases, if the reward depends only on the school's location, the RL agent receives positive reinforcement without actually exerting guidance control, failing to learn appropriate guiding behaviors.

To address this, we define $r_{\beta}$ as a weighted sum of cohesion and directional guidance using a hyperparameter $\beta \in [0, 1]$:
\begin{equation} \label{eq:r_beta}
  r_{\beta} = \beta r_{\mathrm{school}} + (1 - \beta) r_{\mathrm{direction}},
\end{equation}
where both reward terms are normalized to the range $[-1, 1]$ to ensure a balanced contribution to the composite reward.

The cohesion term $r_{\mathrm{school}}$ explicitly evaluates the coupling between the live fish and the virtual agents. It rewards the agents for maintaining proximity to the real individuals, thereby facilitating social influence and preventing the agents from receiving rewards without exerting guidance control. Our formulation calculates the distance to the nearest virtual agent for each real individual:
\begin{equation} \label{eq:r_school}
  r_{\mathrm{school}} = 1 - \frac{\sqrt{2} }{N_r} \sum_{j=1}^{N_r} \min_{i \in \{1, \dots, N_v\}} d(\bm{x}^{(r)}_j, \bm{x}^{(v)}_i),
\end{equation}
where $d(\bm{x}^{(r)}_j, \bm{x}^{(v)}_i)$ represents the Euclidean distance between real fish $j \in \{1, \dots, N_r\}$ and virtual agent $i \in \{1, \dots, N_v\}$. This term encourages virtual agents to maintain proximity to the real fish while simultaneously allowing multiple agents to distribute themselves to manage fragmented subgroups, providing robustness against school splitting.

\begin{figure}[tbp]
  \centering
  \includegraphics[width=0.9\linewidth]{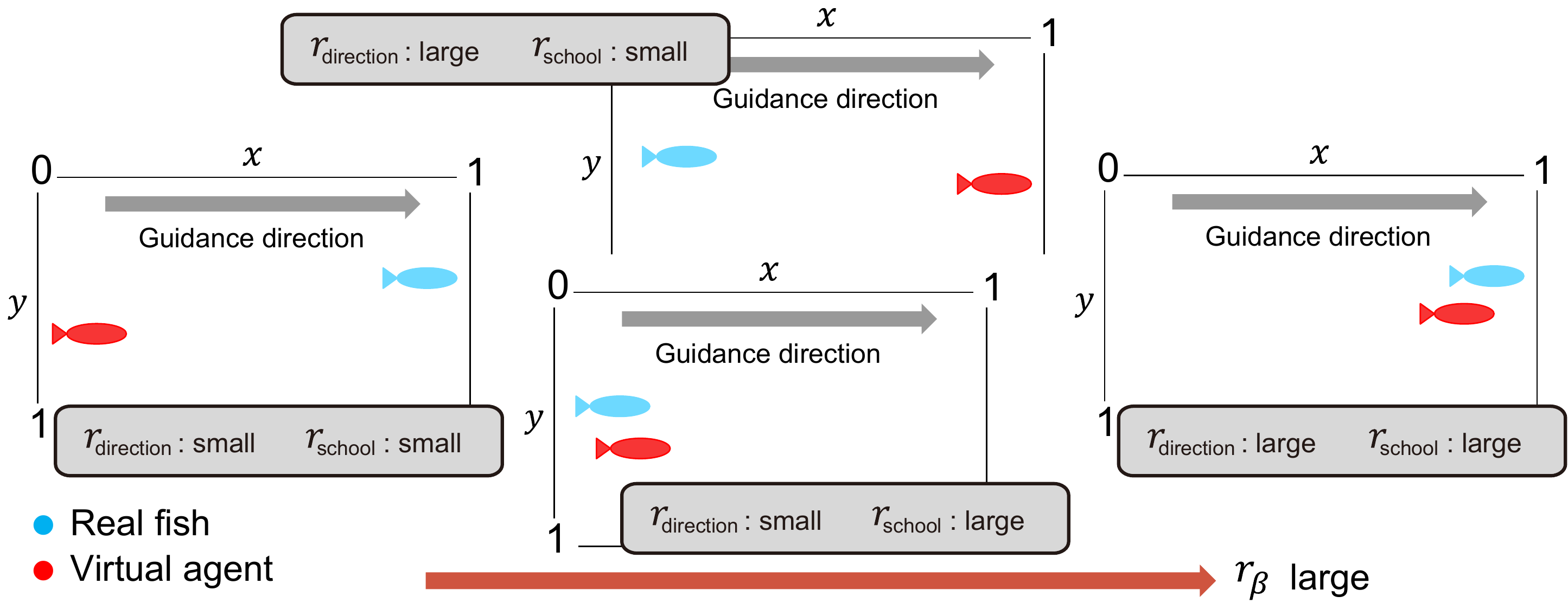}
  \caption{Conceptual diagram of the multi-objective reward design. The cohesion term $r_{\mathrm{school}}$ encourages the virtual agents to maintain proximity to the real fish, while $r_{\mathrm{direction}}$ rewards the progress of the virtual agents toward the target area.}
  \label{fig:reward_conceptual}
\end{figure}

The directional guidance term $r_{\mathrm{direction}}$ evaluates the progress of the virtual agents toward the target end of the tank:
\begin{equation} \label{eq:r_direction}
  r_{\mathrm{direction}} = 1 - 2 | c_x^{(v)} - x_{\mathrm{target\text{-}end}} |
  = \begin{cases}
      2 c_x^{(v)} - 1 & (\text{rightward}) \\
      1 - 2 c_x^{(v)} & (\text{leftward})
    \end{cases},
\end{equation}
where $c_x^{(v)}$ is the horizontal coordinate of the virtual agents' centroid.
The overall structure of this multi-objective reward is illustrated in Fig.~\ref{fig:reward_conceptual}.

The above formulation is presented in a general form applicable to multi-agent settings. During policy learning in simulation, however, we consider the single-agent case ($N_v = 1$), in which the reward terms are reduced to forms defined with respect to the centroid of the real fish school.

\subsection{Agent training in simulation}

\subsubsection{Motivation for simulation-based training}
The primary motivation for employing a simulation environment is to acquire an optimal control policy through pre-training before deploying it in a physical environment. Reinforcement learning typically requires a massive number of interactions to converge, on the order of $10^6$ steps in this study, which is impractical to perform with live animals due to time constraints and the need to ensure animal welfare. Unlike previous work~\cite{Nishii18-26}, our policy was trained entirely in simulation and subsequently deployed in the physical environment without further weight updates or online learning. This approach allowed us to rigorously evaluate the robustness of the policy and its capacity for zero-shot transfer from a virtual model to real biological systems.

\subsubsection{Simulation setup}
In the simulation phase, we utilized the Global mode as defined in Section~\ref{sec:state_action}. 
Individual real fish were not modeled explicitly; instead, the group was represented solely by its centroid.
Specifically, the environment consists of one virtual agent and a simulated school centroid, with their states represented by the normalized coordinates $\bm{x}^{(v)}$ and $\bm{c}^{(r)}$, respectively. 
Under this configuration, the reward $r_{\mathrm{school}}$ (Eq.~\ref{eq:r_school}) simplifies to $1 - \sqrt{2}\, d(\bm{c}^{(r)}, \bm{x}^{(v)})$, and $r_{\mathrm{direction}}$ (Eq.~\ref{eq:r_direction}) is calculated based on the $x$-coordinate of a single virtual agent.

To approximate the continuous dynamics of the agents and the school centroid, the underlying simulation state is updated at a simulation time step of \qty{0.1}{\s}, while the agent selects a discrete action every \qty{1.0}{\s} of simulation time. 
This multi-rate update scheme allows the virtual agent to interact with the collective at a lower update frequency that better matches the characteristic behavioral timescale of the fish, while maintaining the fine temporal resolution required for smooth motion and stable numerical integration of the first-order lag dynamics.
The parameters used for the simulated virtual-agent and school dynamics are listed in Supplementary Table~\ref{tab:supp_dynamics_params}, and the PPO hyperparameters used in the simulation-based policy training are listed in Supplementary Table~\ref{tab:supp_ppo_params}.

\subsubsection{Behavioral model for simulated real fish}
The movement of the simulated fish school centroid $\bm{c}^{(r)}$ was governed by the stochastic behavioral model proposed in \cite{Nishii18-26}. This model assumes that the school's motion consists of a sequence of discrete linear trajectories, reflecting the burst-and-coast swimming style of rummy-nose tetras. 
Under this framework, the school is assumed to behave as a highly cohesive unit, where individual movements are sufficiently synchronized to be effectively represented by their collective centroid.
To account for the non-deterministic nature of social interactions, we introduce an ignoring probability $p$, defined as the probability that the school ignores the virtual agent, following the approach in \cite{Nishii18-26}. Incorporating this probabilistic element encourages the RL agent to develop robust policies that do not rely on guaranteed, deterministic reactions from the fish.

Each phase duration $\Delta t \in (0, \Delta t_{\max}]$ is uniformly sampled. During each phase, the velocity follows a first-order lag system:
\begin{equation} \label{eq:real_move_sim}
  \frac{\dd \bm{c}^{(r)}}{\dd t} = \frac{1}{\tau^{(r)}}(\bm{c}^{(r)}_{\mathrm{target}} - \bm{c}^{(r)}),
\end{equation}
where $\bm{c}^{(r)}_{\mathrm{target}}$ is the target coordinate updated at the beginning of each phase based on the distance $d(\bm{c}^{(r)}, \bm{x}^{(v)})$ and the ignoring probability $p$, as follows:
\begin{itemize}
    \item \textbf{Reaction case}: If $d(\bm{c}^{(r)}, \bm{x}^{(v)}) \le \theta$ (within the interaction range), the school reacts to the agent with probability $1-p$, setting $\bm{c}^{(r)}_{\mathrm{target}} = \bm{x}^{(v)}$.
    \item \textbf{Spontaneous movement}: If $d(\bm{c}^{(r)}, \bm{x}^{(v)}) > \theta$, or with probability $p$ even within the interaction range, the target is updated by a random displacement:
    \begin{equation} \label{eq:target_random_sim}
        \bm{c}^{(r)}_{\mathrm{target}} = \bm{c}^{(r)} + \left(\begin{array}{c} \delta x \\ \delta y \end{array}\right),
    \end{equation}
    where $\delta x$ and $\delta y$ are sampled uniformly from $[-\delta x_{\max}, \delta x_{\max}]$ and $[-\delta y_{\max}, \delta y_{\max}]$, respectively.
\end{itemize}

\subsubsection{Evaluation procedure}
The performance of the acquired policy is evaluated over a validation period of $T' = 5000$ steps with the learned policy parameters (network weights) held fixed.
To ensure a consistent comparison between agents trained with different values of $\beta$, the evaluation metric $R$ is defined as the time-average of the baseline reward $r_{\mathrm{base}}$ (Eq.~(\ref{eq:r_base})) over an evaluation period:
\begin{equation} \label{eq:eval_R_metric}
    R = \frac{1}{T'} \sum_{t=1}^{T'} r_{\mathrm{base}}(t)
\end{equation}
Because $r_{\mathrm{base}}$ depends only on $c_x^{(r)}$, the horizontal coordinate of the real school centroid, it provides an objective benchmark for identifying the policy that yields the most effective guidance behavior.

\subsection{Physical experiment protocols and evaluation}
\label{sec:physical_experiments}

\subsubsection{Experimental protocol}
\label{sec:experimental_protocol}

Based on the swimming speed of rummy-nose tetras, the duration of each control step for the virtual agents was set to \qty{1.2}{\s} for all physical experiments. 
The parameters of the virtual-agent dynamics used in the physical experiments are summarized in Supplementary Table~\ref{tab:supp_dynamics_params}.

Each experimental trial consisted of 900 steps (approximately 18 minutes), with the target direction initially set rightward and switched every 90 steps to evaluate the adaptive response of the school. 
Four trials were conducted for each experimental configuration at various times and on multiple dates.
The order of the experimental conditions was varied across days by cyclically rotating the starting condition. Experiment~A and the $N_r = 8$ conditions of Experiment~B were each completed over three days, whereas the $N_r = 5$ conditions of Experiment~B were completed over four days due to a technical issue.

The fish were kept in separate holding tanks and moved to the experimental arena only for the duration of the trials. 
The room temperature was maintained at \qty{25}{\degreeCelsius}, and the water temperature was approximately \qty{27}{\degreeCelsius}. The fish were fed once daily in the morning.

For each trial, individuals were randomly selected from the holding population: three fish from a population of five for Experiment~A, and five or eight fish from a separate population of twelve for Experiment~B.
To perform the guidance tasks in stable, still-water conditions, the water circulation system was temporarily suspended during all experiments.
To ensure stable behavioral states, the fish were introduced to the experimental tank at least 90 minutes before the start of the trials. When switching between different experimental conditions, a 30-minute interval was maintained to minimize the carryover effects of the previous stimuli.

\subsubsection{Experimental design}
\label{sec:experimental_design}
The physical experiments were conducted in two stages, described below.
\begin{itemize}
    \item \textbf{Experiment~A}: This stage aimed to identify the experimental conditions that maximized the stimulus salience of the virtual agents. We systematically tested three background colors (white, gray, and black, whose pixel values were 255, 123, and 0 for all RGB channels, respectively) and three fish-image sizes (small, medium, and large, which were approximately $0.6\times$, $1.0\times$, and $1.5\times$ the size of real fish, respectively).
    In the background-color experiment, the fish-image size was fixed at the large setting, whereas in the fish-image-size experiment, the background color was fixed at white; these fixed settings were determined in preliminary trials.
    These trials were performed with a group of three real individuals ($N_r = 3$) using a single fixed-formation virtual agent.
    The fixed-formation agent consisted of four fish images arranged in a diamond-shaped configuration with fixed relative positions and small gaps between adjacent images.
    The agent operated in the Global mode, targeting the centroid of the entire school as described in Section~\ref{sec:state_action}.
    \item \textbf{Experiment~B}: Using the visual parameters (background color and fish-image size) identified in Experiment~A, we evaluated the guidance performance across different group sizes ($N_r = 5, 8$) and agent configurations. We compared the fixed-formation configuration used in Experiment~A, treated as a single-unit agent ($N_v = 1$), with independently controlled agents ($N_v = 2, 3$). 
    In contrast to the fixed-formation agent, the independent agents were each rendered as a single fish image and operated in the Cluster-assignment mode, where each agent targeted a localized subgroup as defined in Section~\ref{sec:state_action}.
\end{itemize}

\begin{figure}[tbp]
  \centering
  \begin{subfigure}[b]{0.45\textwidth}
    \centering
    \includegraphics[width=\textwidth]{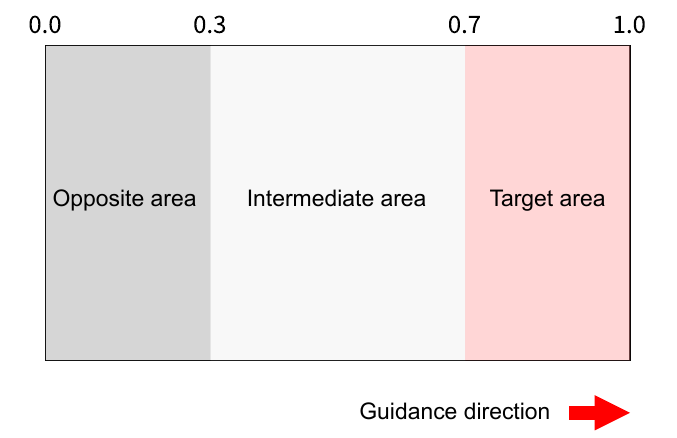}
    \caption{Rightward guidance}
    \label{fig:area_right}
  \end{subfigure}
  \hfill
  \begin{subfigure}[b]{0.45\textwidth}
    \centering
    \includegraphics[width=\textwidth]{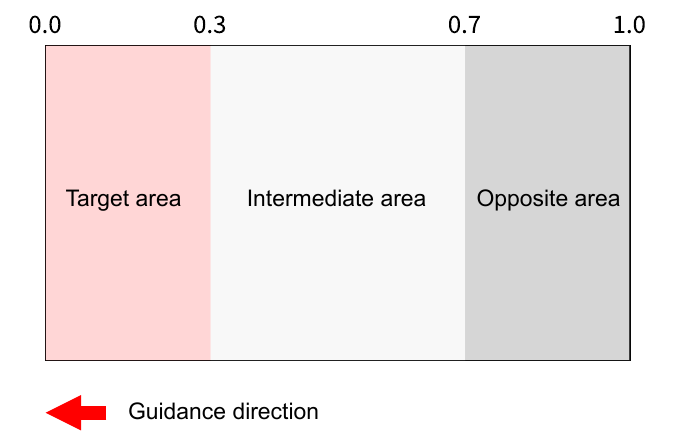}
    \caption{Leftward guidance}
    \label{fig:area_left}
  \end{subfigure}
  \caption{Definition of evaluation areas for guidance tasks. The target area is defined as the 30\% region from the target end, while the opposite area is the 30\% region from the opposite end.}
  \label{fig:eval_areas}
\end{figure}

\subsubsection{Evaluation metrics}
\label{sec:eval_metrics}

The automated fish-detection system (Section~\ref{sec:vision_mapping}) provided the positions of all detected individuals in every camera frame, from which the guidance reference points were computed. 
For offline analysis, fish positions were recorded at each control step (\qty{1.2}{\s}), and only the $N_r$ detections with the highest confidence were retained when more than $N_r$ individuals were detected.

To remove static false detections from fixed artifacts, we additionally pooled the retained positions for each recording date, binned them on a spatial grid (cell size \num{0.02} in normalized coordinates), and discarded detections in any cell whose count reached at least 1\% of that date's total, treating them as undetected.

Control steps in which no individuals were detected were excluded from the analysis. 
Valid steps and detection rates are reported in Supplementary Table~\ref{tab:supp_trials}.

Guidance efficacy was evaluated using the following metrics:
\begin{enumerate}
    \item \textbf{Area occupancy ratio}: The tank was divided into three functional zones according to horizontal position (Fig.~\ref{fig:eval_areas}): the target area (the 30\% region nearest to the target end), the opposite area (the 30\% region at the opposite end), and the intermediate area (the central 40\%). 
    The percentage of time spent by the detected individuals in each zone was calculated for each trial and then averaged across the four trials for each condition.
    \item \textbf{Guidance metric $R$}: As the primary quantitative measure of guidance efficacy, we used the scalar metric $R \in [-1, 1]$ (Eq.~\ref{eq:eval_R_metric}). This metric was computed as the time average of the baseline reward $r_\text{base} \in [-1, 1]$ (Eq.~\ref{eq:r_base}), evaluated from the horizontal coordinate of the school centroid. As with the area occupancy ratio, $R$ was calculated for each trial and then averaged across the four trials for each condition. 
    For the statistical analysis (Section~\ref{sec:statistical_analysis}), the same time average was computed over each 90-step sub-interval, yielding the sub-interval metric $R_s$ for the $s$-th sub-interval.
    \item \textbf{Directional distribution and Bhattacharyya distance}: Positional 20-bin histograms of the horizontal coordinate of the school centroid were generated separately for the leftward and rightward guidance periods, with data pooled across the four trials. The separability of the two distributions in each condition was summarized using the Bhattacharyya distance, with larger values indicating greater separation and thus more effective directional guidance.
    \item \textbf{Sub-interval distribution}: To assess the temporal stability of guidance for representative configurations, the distribution of individual horizontal positions within each 90-step sub-interval was visualized using box plots.
\end{enumerate}

\subsubsection{Statistical analysis}
\label{sec:statistical_analysis}

Statistical tests of the sub-interval guidance metric ($R_s$) were performed using linear mixed-effects models (LMMs), with the trial as the unit of replication. 
Because the fish were drawn anew from a small holding population for each trial (Section~\ref{sec:experimental_protocol}) and their individual identities were not recorded, individual fish could not be included as a grouping factor; the resulting non-independence among trials is considered in Section~\ref{sec:limitation_asymmetry}.

For Experiment~A, separate models were fitted for background color and fish-image size, with the corresponding experimental factor included as a fixed effect. For Experiment~B, group size, agent configuration, and their interaction were included as fixed effects. Target direction (leftward or rightward) was included as a fixed effect in all models to account for an overall directional bias.

In both experiments, the random-effects structure comprised a trial-specific random intercept and an uncorrelated trial-specific random slope for target direction. For the random slope, target direction was coded as a centered numeric variable. The fixed effect of target direction accounted for the overall left--right bias, whereas the random slope allowed this directional bias to vary among trials.

Statistical inference comprised three steps: (1) fixed effects were tested using Type~III $F$-tests with the Kenward--Roger approximation for the denominator degrees of freedom; (2) for each condition, guidance was assessed by testing the estimated marginal mean of $R$, averaged over target direction, against zero; and (3) conditions were compared using pairwise contrasts, with $p$-values adjusted using the Holm method. 

All statistical tests were performed in R using the \texttt{lme4}, \texttt{lmerTest}, \texttt{pbkrtest}, and \texttt{emmeans} packages.
Area occupancy ratios and Bhattacharyya distances were treated as descriptive measures and were not subjected to inferential testing.

\section{Results}

\subsection{Policy optimization through simulation}
Prior to the physical experiments, we evaluated the effectiveness of the reinforcement learning framework in a simulation environment. The primary objectives were to determine the optimal weight $\beta$ for the composite reward function $r_{\beta}$ (Eq.~(\ref{eq:r_beta})) and to ensure that the acquired policy remains robust across various levels of stochasticity in fish behavior, represented by the ignoring probability $p$.
To account for the stochastic nature of the learning process, we performed 10 independent training runs for each parameter combination and averaged the evaluation metric $R$ across these runs.

\begin{figure}[tbp]
  \centering
  \includegraphics[width=0.90\linewidth]{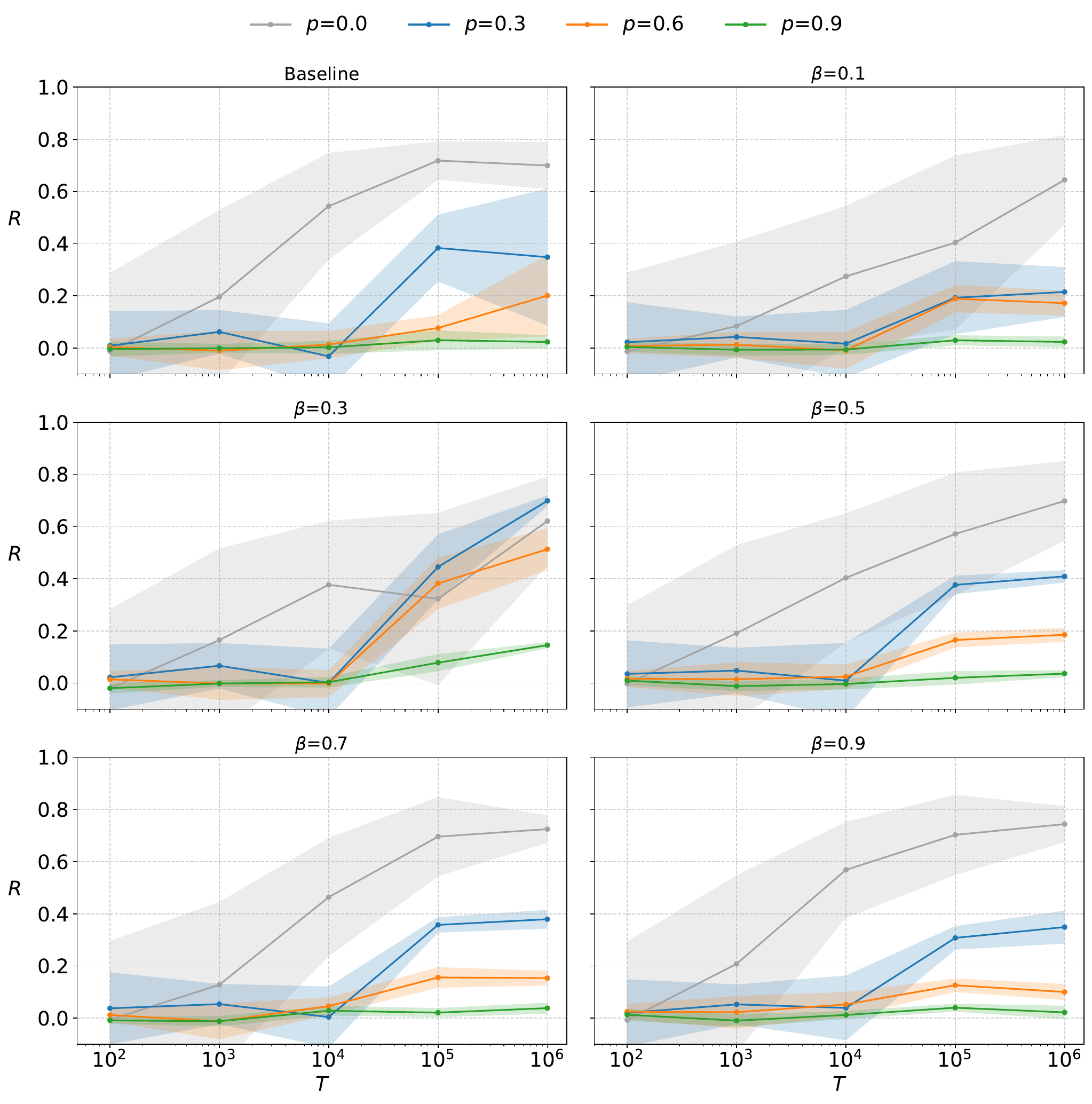}
  \caption{Learning curves showing the transition of the evaluation metric $R$ across different training steps $T$ and reward weights $\beta$. Each data point represents the mean of $R$ over 10 independent training runs for each corresponding parameter combination, and the shaded regions represent the 95\% confidence intervals ($\pm 1.96$ standard errors). The baseline represents the policy trained using only the horizontal coordinate of the school's centroid ($r_{\mathrm{base}}$). Each plot compares different ignoring probabilities $p$ for the simulated fish.}
  \label{fig:simulation_result}
\end{figure}

Figure~\ref{fig:simulation_result} illustrates the transition of the evaluation metric $R$ as a function of training steps $T$. Overall, the performance generally improved as $T$ increased across all parameter combinations. When comparing different reward configurations, we observed that policies trained with $\beta = 0.1, 0.5, 0.7,$ and $0.9$ yielded performance levels comparable to the baseline reward $r_{\mathrm{base}}$. However, the policy trained with $\beta = 0.3$ showed a higher mean $R$ than the baseline at $T = 10^6$ steps for all values of $p$ except $p=0$. This $p=0$ case represents an idealized scenario in which the fish always react to the agent; it is included as a reference point for comparison.

The results also confirmed that higher values of the ignoring probability $p$ generally led to lower $R$, reflecting the increased difficulty of the guidance task when the school frequently ignores the virtual agent. Nevertheless, the composite reward function $r_{\beta}$ with an appropriate value of $\beta$ successfully facilitated stable policy acquisition even under high-stochasticity conditions ($p \geq 0.6$). 
Based on preliminary observations of the live fish's responses to nearby individuals, $p = 0.6$ was chosen as a representative level of response stochasticity. We then adopted the policy trained with $T = 10^6$, $p = 0.6$, and $\beta = 0.3$ as the autonomous agent controller for all subsequent physical experiments described in Section~\ref{sec:physical_experiments}.

\subsection{Experiment~A: Background color and fish-image size}

In Experiment~A, we systematically evaluated stimulus salience by varying the background color and fish-image size to maximize the responsiveness of the live fish. As described in Section~\ref{sec:experimental_design}, these trials were performed with a group of three individuals ($N_r = 3$) using a single fixed-formation agent operating in the Global mode.

Table~\ref{tab:experiment_a_results} summarizes the guidance performance for each condition. Per-trial values underlying this summary are provided in Supplementary Table~\ref{tab:supp_trials}.
Detailed statistical results for the physical experiments, corresponding to the analyses described in Section~\ref{sec:statistical_analysis} (Type III tests, estimated marginal means, and pairwise contrasts), are reported in Supplementary Tables~\ref{tab:supp_ftests}--\ref{tab:supp_pairwise}.

\begin{figure}[tb]
  \centering
  \begin{subfigure}[b]{0.62\textwidth}
    \centering
    \includegraphics[width=\textwidth]{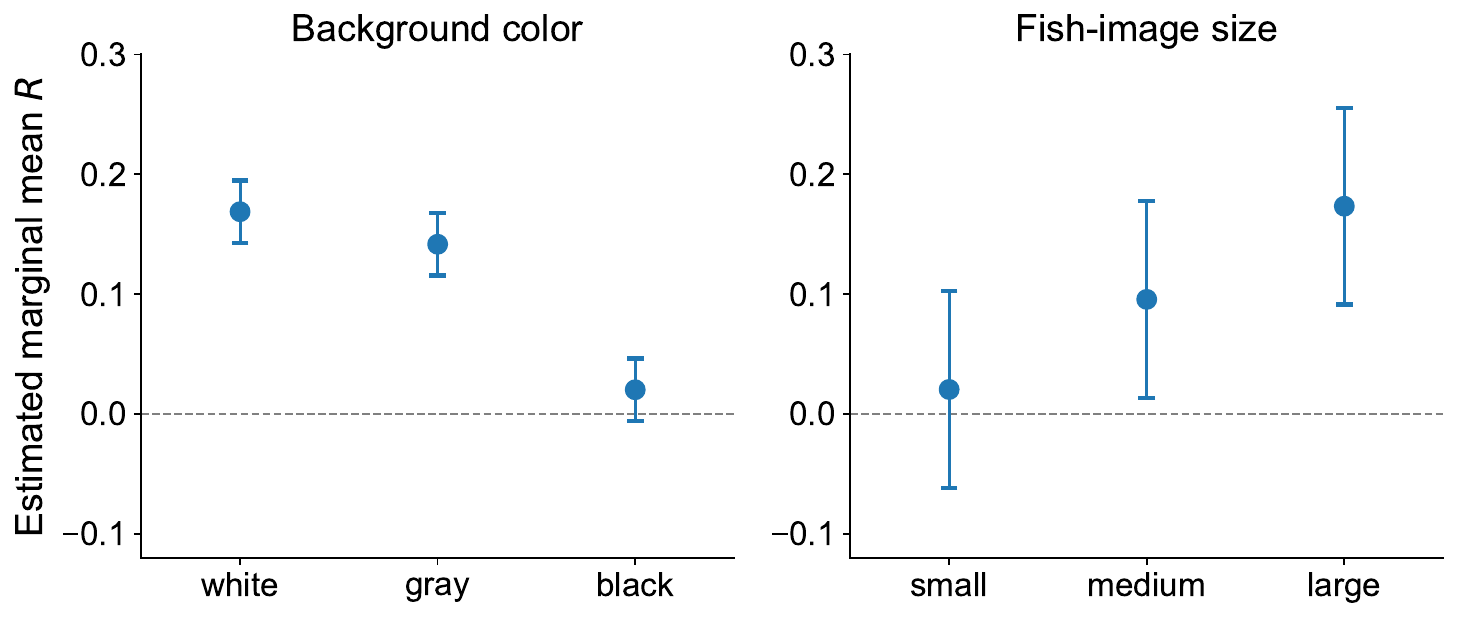}
    \caption{Experiment~A (left: background color, right: fish-image size)}
    \label{fig:emm_experiment_a}
  \end{subfigure}
  \hfill
  \begin{subfigure}[b]{0.34\textwidth}
    \centering
    \includegraphics[width=\textwidth]{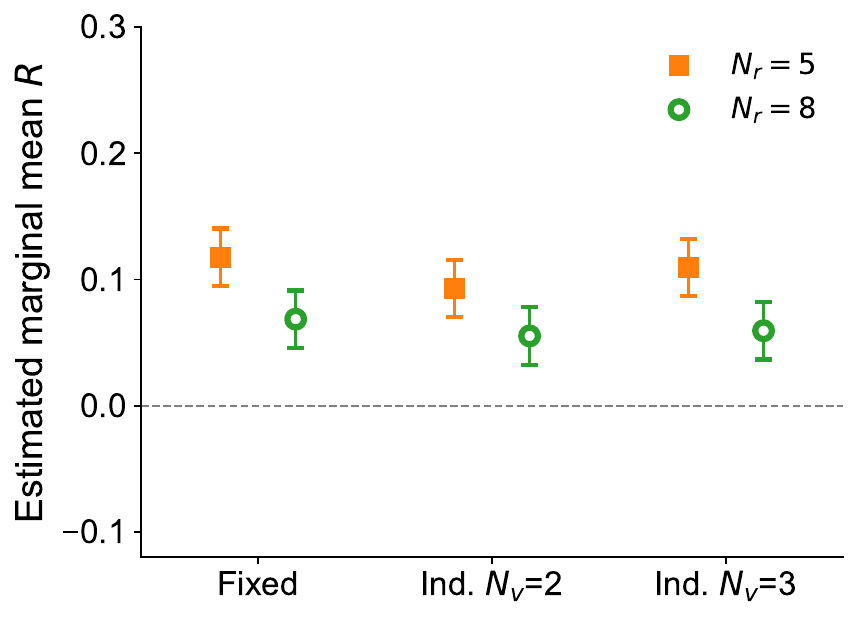}
    \caption{Experiment~B}
    \label{fig:emm_experiment_b}
  \end{subfigure}
  \caption{Estimated marginal means of the guidance metric $R$
    (Eq.~\ref{eq:eval_R_metric}), averaged over target direction, with 95\%
    confidence intervals (Kenward--Roger) from the linear mixed models
    (Section~\ref{sec:statistical_analysis}). The dashed line marks $R=0$ (no
    directional bias). (a)~Experiment~A ($N_r=3$): background color (left) and
    fish-image size (right). (b)~Experiment~B: filled squares indicate
    $N_r=5$ and open circles indicate $N_r=8$.}
  \label{fig:emm_summary}
\end{figure}

\begin{figure}[H]
  \centering
  \begin{subfigure}[b]{0.48\textwidth}
    \centering
    \includegraphics[width=0.95\textwidth]{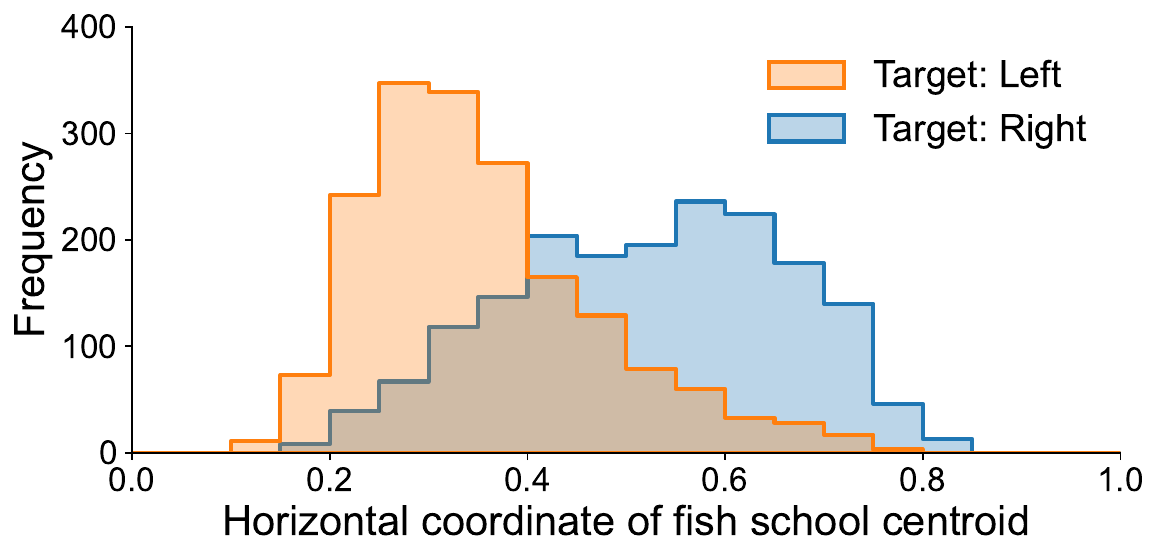}
    \centerline{\small Background: white (BD=0.201)}\vspace{2ex}
    \includegraphics[width=0.95\textwidth]{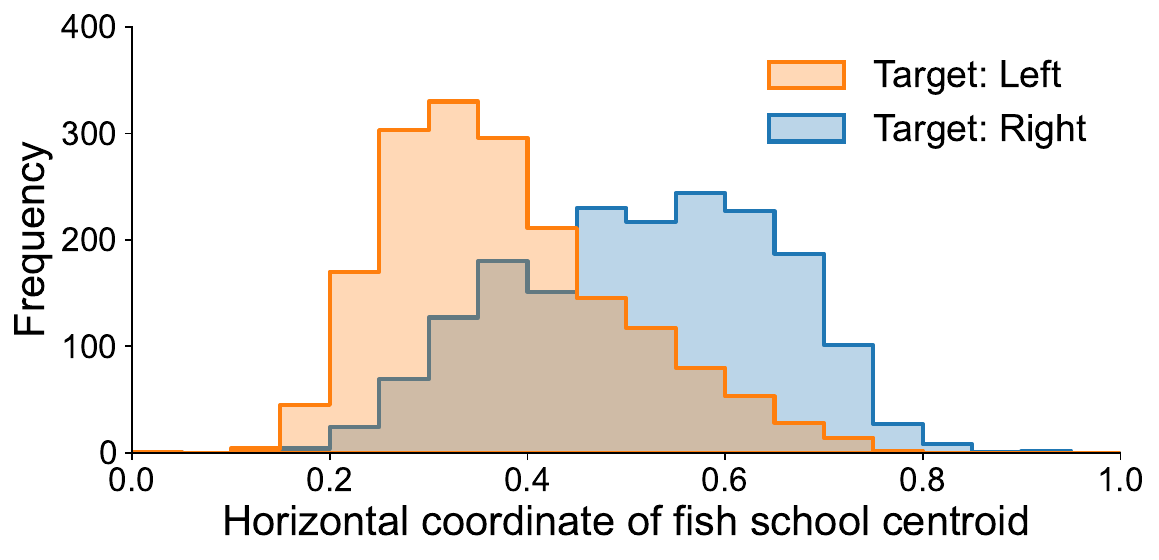}
    \centerline{\small Background: gray (BD=0.152)}\vspace{2ex}
    \includegraphics[width=0.95\textwidth]{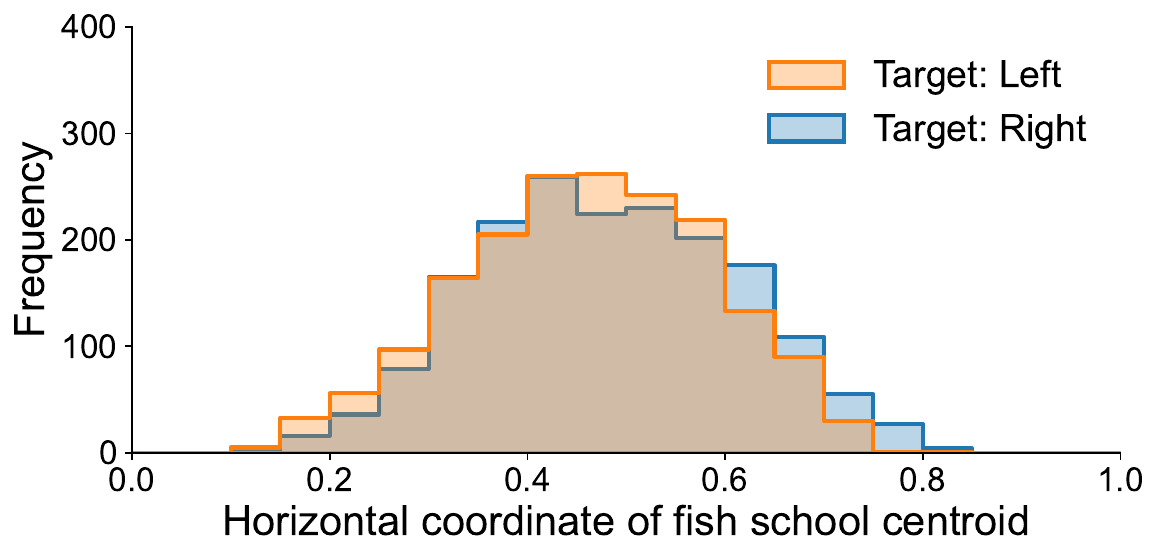}
    \centerline{\small Background: black (BD=0.010)}
    \caption{Effect of background color}
    \label{fig:hist_bg}
  \end{subfigure}
  \hfill
  \begin{subfigure}[b]{0.48\textwidth}
    \centering
    \includegraphics[width=0.95\textwidth]{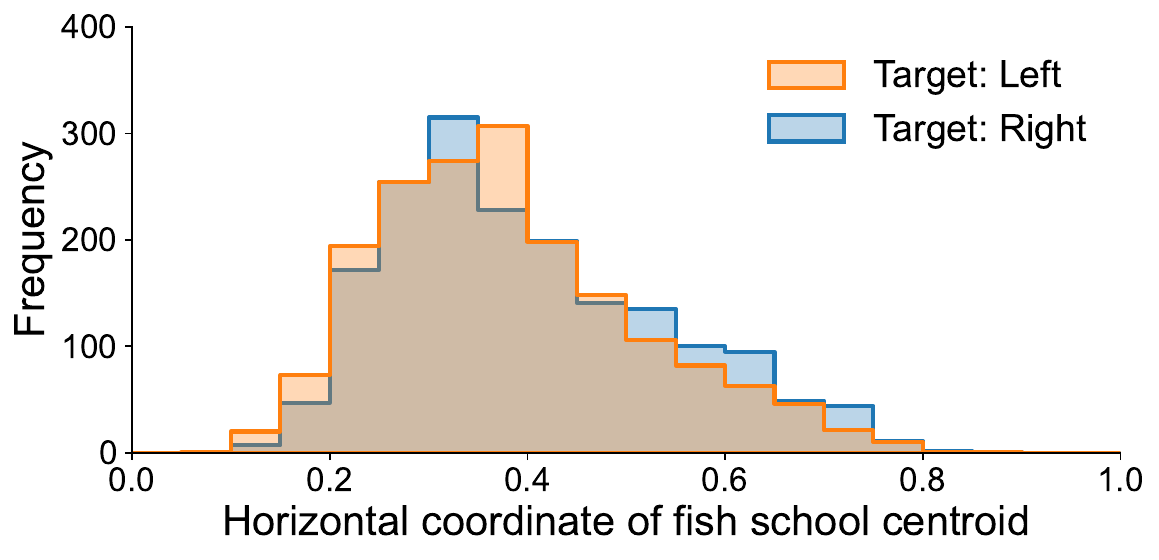}
    \centerline{\small Fish-image size: small (BD=0.007)}\vspace{2ex}
    \includegraphics[width=0.95\textwidth]{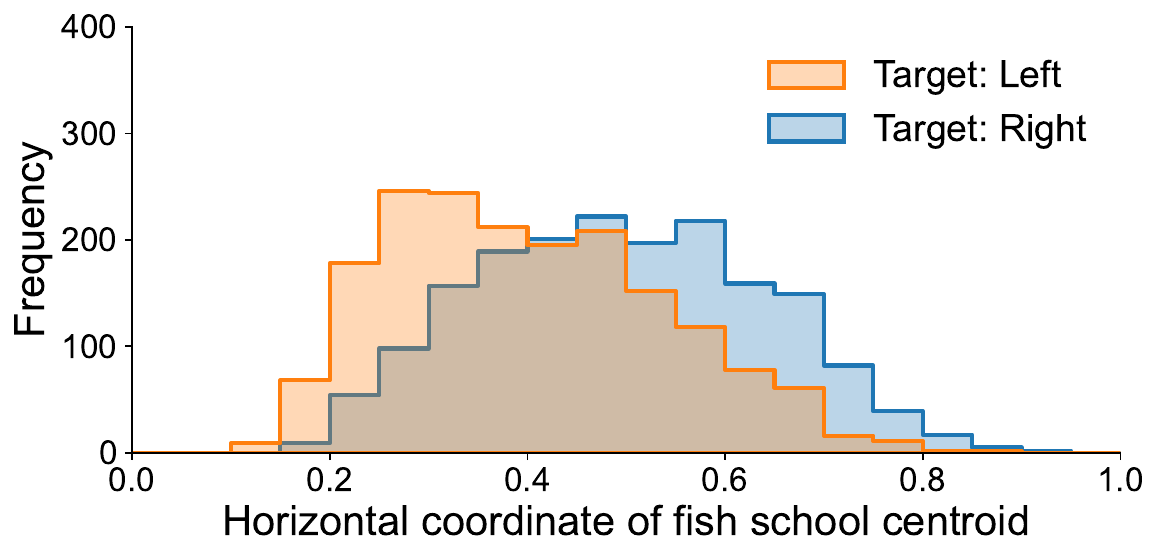}
    \centerline{\small Fish-image size: medium (BD=0.061)}\vspace{2ex}
    \includegraphics[width=0.95\textwidth]{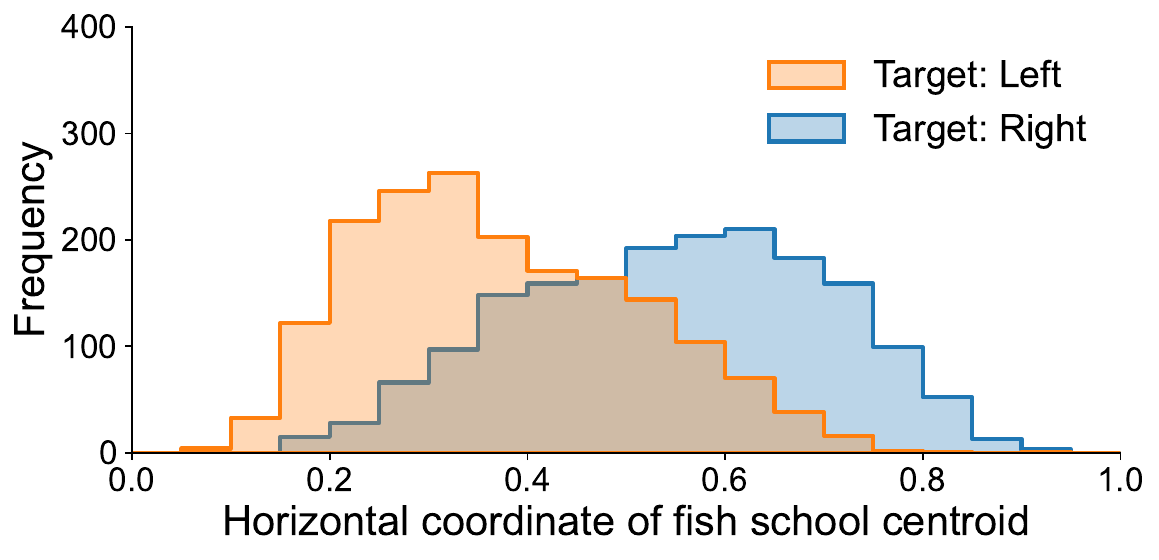}
    \centerline{\small Fish-image size: large (BD=0.180)}
    \caption{Effect of fish-image size}
    \label{fig:hist_size}
  \end{subfigure}
  \caption{Positional histograms of the school's centroid for the visual parameters tested in Experiment~A. (a) Background color comparison. (b) Fish-image size comparison (small: $0.6\times$, medium: $1.0\times$, large: $1.5\times$). Each plot shows the overlay of leftward and rightward guidance results.}
  \label{fig:hist_experiment_a}
\end{figure}

\paragraph{Background color.} Under the white and gray backgrounds, the detected individuals spent
far more time in the target area than in the opposite area ($29.4\%$ vs.\ $7.4\%$ and $25.4\%$ vs.\ $7.1\%$, respectively), whereas the occupancy ratios of the two areas were comparable under the black background ($12.5\%$ vs.\ $9.1\%$), indicating a loss of directional bias. This pattern was also reflected in the guidance metric: background color had a significant effect on the metric $R$ (LMM, $F_{2,9}=47.0$, $p<0.001$), and the estimated marginal mean of $R$ was significantly greater than zero under white (95\% CI, $0.14$--$0.19$) and gray ($0.12$--$0.17$), but not under black ($-0.01$--$0.05$) (Fig.~\ref{fig:emm_experiment_a} left).
Both white and gray outperformed black (Holm-adjusted $p<0.001$ for both), whereas white and gray did not differ significantly ($p=0.130$).
The positional histograms (Fig.~\ref{fig:hist_bg}) corroborate this result: the centroid distribution shifted clearly toward the target direction, with the shift being most pronounced during leftward guidance under white and gray backgrounds, but remained largely centered under black. Correspondingly, the Bhattacharyya distance between the leftward and rightward distributions was high under white ($0.201$) and gray ($0.152$), and negligible under black ($0.010$).

\begin{table}[H]
  \centering
  \caption{Summary of guidance performance in Experiment~A. Area occupancy ratios are the percentage of time individuals spent in each zone, and $R$ is the guidance evaluation metric (Eq.~\ref{eq:eval_R_metric}), with larger $R$ indicating stronger guidance toward the target. Values are mean $\pm$ SD over four trials ($n=4$).}
  \label{tab:experiment_a_results}
  \begin{tabular}{ll r@{\,$\pm$\,}l r@{\,$\pm$\,}l r@{\,$\pm$\,}l}
    \hline
    Parameter & Condition
      & \multicolumn{2}{c}{\begin{tabular}[c]{@{}c@{}}Target\\area (\%)\end{tabular}}
      & \multicolumn{2}{c}{\begin{tabular}[c]{@{}c@{}}Opposite\\area (\%)\end{tabular}}
      & \multicolumn{2}{c}{$R$} \\ \hline
    Background color & white          & 29.4 & 3.0  & 7.4  & 2.4 & 0.17 & 0.02 \\
                     & gray           & 25.4 & 2.8  & 7.1  & 1.9 & 0.14 & 0.01 \\
                     & black          & 12.5 & 7.9  & 9.1  & 5.6 & 0.02 & 0.02 \\ \hline
    Fish-image size  & small ($0.6\times$)  & 20.2 & 3.3  & 17.9 & 4.3 & 0.02 & 0.04 \\
                     & medium ($1.0\times$) & 24.0 & 9.2  & 11.7 & 1.6 & 0.10 & 0.07 \\
                     & large ($1.5\times$)  & 31.3 & 11.3 & 8.4  & 1.9 & 0.17 & 0.10 \\ \hline
  \end{tabular}
\end{table}

\paragraph{Fish-image size.} Regarding fish-image size, the target-area occupancy increased with image size ($20.2\%$, $24.0\%$, and $31.3\%$ for the small, medium, and large sizes, respectively), while the opposite-area occupancy decreased ($17.9\%$, $11.7\%$, and $8.4\%$). The LMM indicated a significant effect of fish-image size ($F_{2,9}=4.45$, $p=0.045$). The estimated marginal mean of $R$ was significantly greater than zero for the large (95\% CI, $0.09$--$0.26$) and medium ($0.01$--$0.18$) sizes, but not for the small size ($-0.06$--$0.10$), indicating that the smallest images did not produce a detectable guidance effect (Fig.~\ref{fig:emm_experiment_a} right). The large--small contrast was significant ($p=0.046$), whereas the large--medium and medium--small contrasts were not significant (both $p=0.33$).
Consistently, the histograms (Fig.~\ref{fig:hist_size}) show a clearer target-directed positional bias for the large size, with the Bhattacharyya distance increasing with image size ($0.007$, $0.061$, and $0.180$) for the small, medium, and large sizes, respectively.

Overall, the results of Experiment~A demonstrate that the white background and the large fish-image size produced the highest behavioral responsiveness among the tested conditions in our physical environment.
Consequently, this combination of visual parameters was adopted and kept fixed for all subsequent trials in Experiment~B.

\subsection{Experiment~B: Agent configuration and group sizes}

In Experiment~B, we evaluated the guidance performance focusing on the influence of agent configurations and group sizes ($N_r = 5, 8$). Based on the findings from Experiment~A, all trials were conducted using the white background and the large fish-image size.

Table~\ref{tab:experiment_b_results} summarizes the guidance performance in Experiment~B (per-trial values are provided in Supplementary Table~\ref{tab:supp_trials}). 
The corresponding detailed statistical results are presented in Supplementary Tables~\ref{tab:supp_ftests}--\ref{tab:supp_pairwise}.

\begin{table}[tb]
  \centering
  \caption{Summary of guidance performance in Experiment~B for group sizes $N_r=5$ and $N_r=8$ across agent configurations. Area occupancy ratios are the percentage of time individuals spent in each zone, and $R$ is the guidance evaluation metric (Eq.~\ref{eq:eval_R_metric}), with larger $R$ indicating stronger guidance. Values are mean $\pm$ SD over four trials ($n=4$).}
  \label{tab:experiment_b_results}
  \begin{tabular}{ll r@{\,$\pm$\,}l r@{\,$\pm$\,}l r@{\,$\pm$\,}l}
    \hline
    $N_r$ & Configuration (Mode)
      & \multicolumn{2}{c}{\begin{tabular}[c]{@{}c@{}}Target\\area (\%)\end{tabular}}
      & \multicolumn{2}{c}{\begin{tabular}[c]{@{}c@{}}Opposite\\area (\%)\end{tabular}}
      & \multicolumn{2}{c}{$R$} \\ \hline
    5 & Fixed (Global)                & 25.7 & 3.0 & 10.2 & 2.4 & 0.12 & 0.03 \\
      & Independent $N_v=2$ (Cluster) & 23.3 & 1.5 & 11.1 & 2.9 & 0.09 & 0.01 \\
      & Independent $N_v=3$ (Cluster) & 25.8 & 3.1 & 10.9 & 2.0 & 0.11 & 0.02 \\ \hline
    8 & Fixed (Global)                & 20.8 & 2.0 & 10.8 & 2.8 & 0.07 & 0.03 \\
      & Independent $N_v=2$ (Cluster) & 20.5 & 2.1 & 12.5 & 1.7 & 0.06 & 0.02 \\
      & Independent $N_v=3$ (Cluster) & 20.9 & 1.8 & 12.9 & 3.2 & 0.06 & 0.02 \\ \hline
  \end{tabular}
\end{table}

\begin{figure}[tb]
  \centering
\begin{subfigure}[b]{0.48\textwidth}
    \centering
    \includegraphics[width=0.95\textwidth]{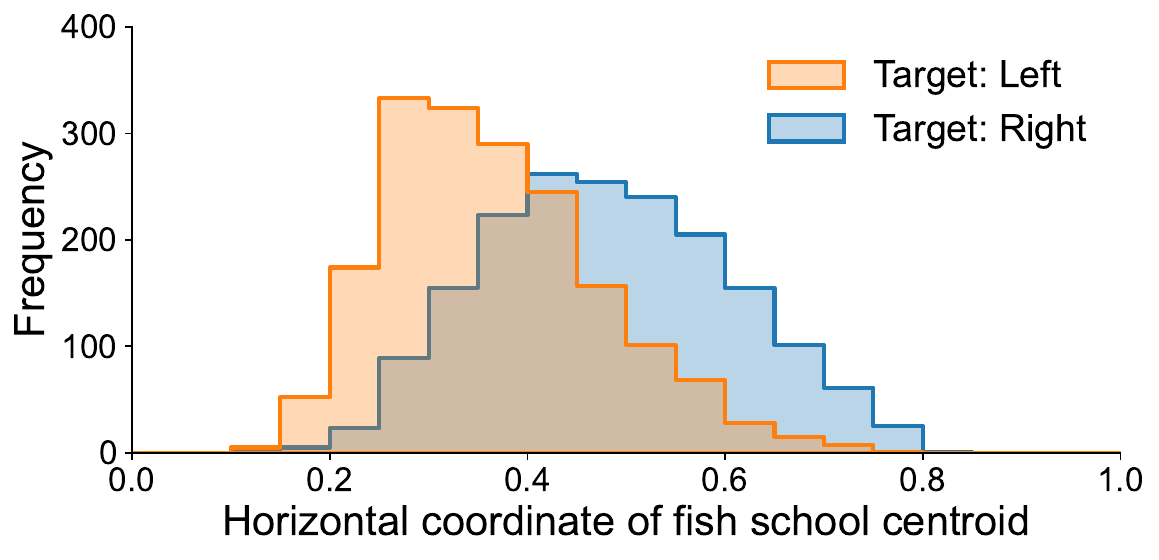}
    \centerline{\small Fixed formation (BD=0.124)}\vspace{2ex}
    \includegraphics[width=0.95\textwidth]{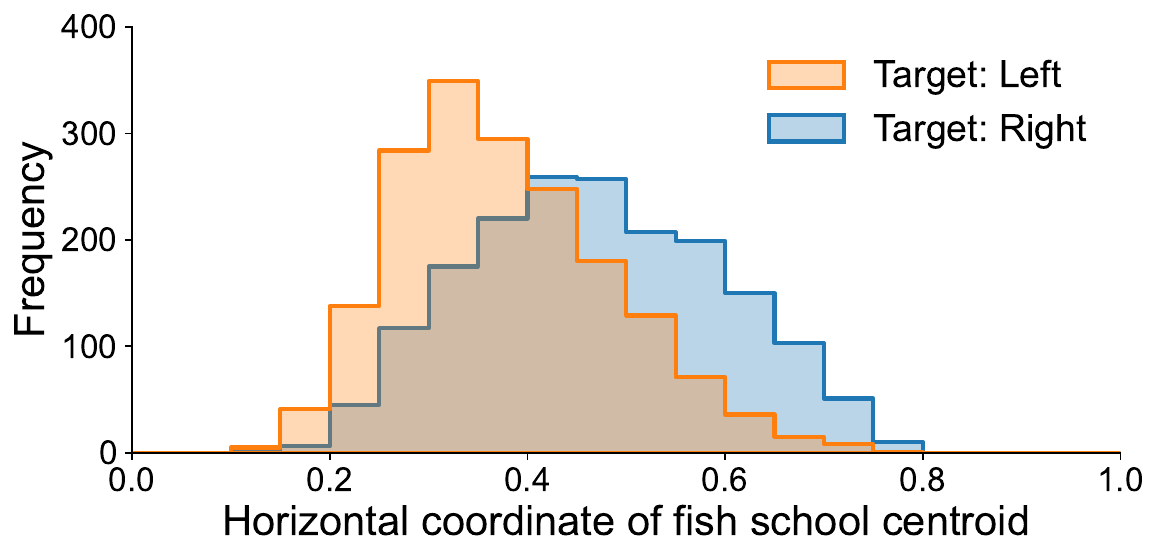}
    \centerline{\small Independent ($N_v=2$) (BD=0.076)}\vspace{2ex}
    \includegraphics[width=0.95\textwidth]{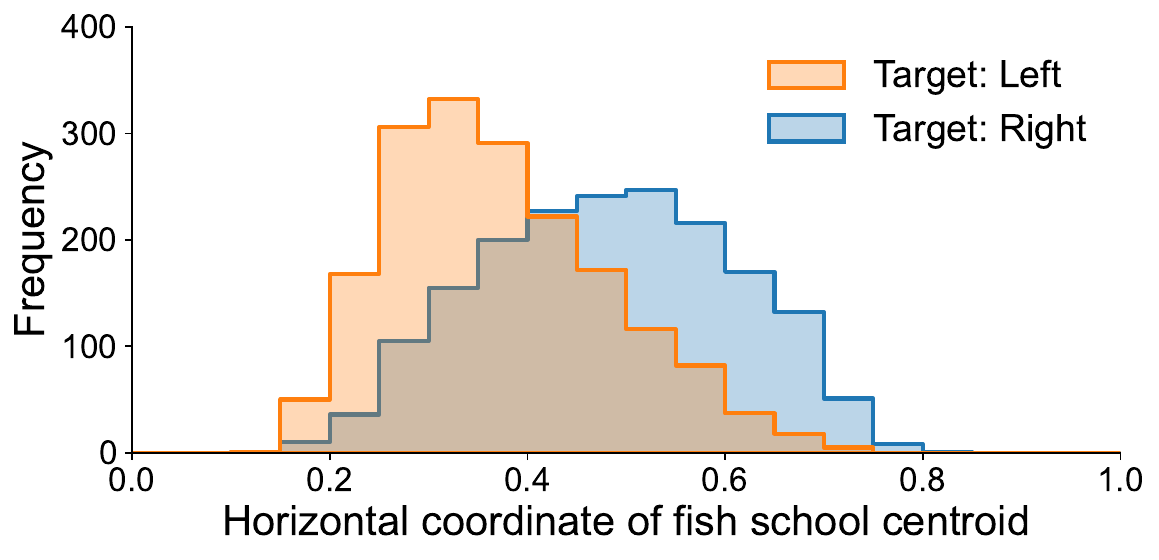}
    \centerline{\small Independent ($N_v=3$) (BD=0.104)}
    \caption{Group size $N_r = 5$}
    \label{fig:hist_r5}
  \end{subfigure}
  \hfill
  \begin{subfigure}[b]{0.48\textwidth}
    \centering
    \includegraphics[width=0.95\textwidth]{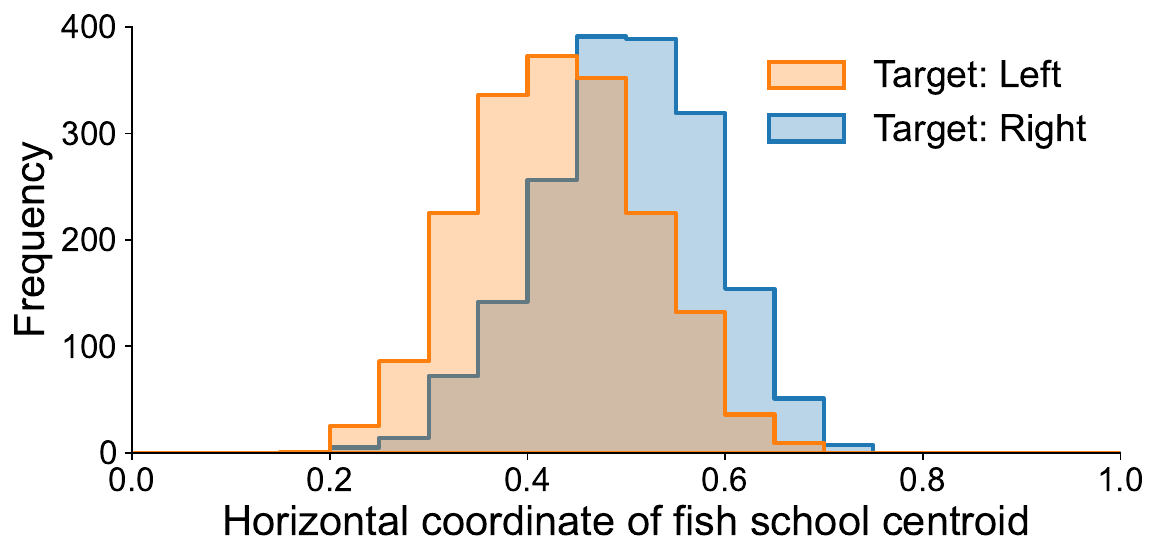}
    \centerline{\small Fixed formation (BD=0.076)}\vspace{2ex}
    \includegraphics[width=0.95\textwidth]{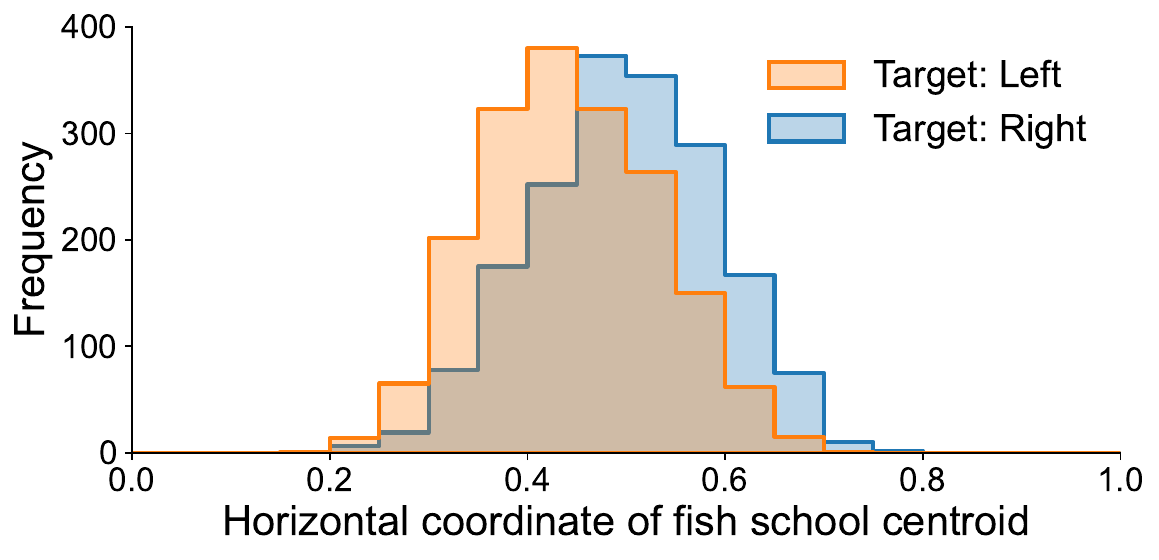}
    \centerline{\small Independent ($N_v=2$) (BD=0.047)}\vspace{2ex}
    \includegraphics[width=0.95\textwidth]{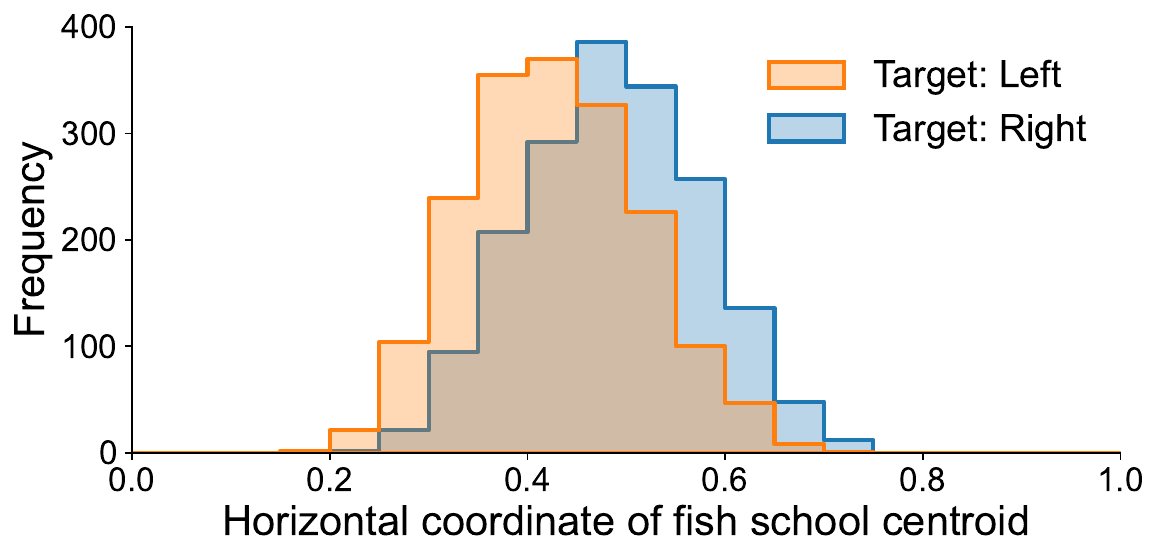}
    \centerline{\small Independent ($N_v=3$) (BD=0.057)}
    \caption{Group size $N_r = 8$}
    \label{fig:hist_r8}
  \end{subfigure}
  \caption{Positional histograms of the school's centroid for Experiment~B. The columns compare the (a) $N_r=5$ and (b) $N_r=8$ conditions. The rows represent the fixed-formation configuration and the independent ($N_v=2, 3$) configurations. Each plot shows the overlay of leftward and rightward guidance results.}
  \label{fig:hist_experiment_b}
\end{figure}

\subsubsection{Guidance efficacy and group-size dependence}

At $N_r=5$, the detected individuals spent $23$--$26\%$ of the time in the target area against $10$--$11\%$ in the opposite area across all three configurations. At $N_r=8$, by contrast, the target-area occupancy fell to about $21\%$ and the opposite-area occupancy rose to $11$--$13\%$, reducing the directional asymmetry. The guidance metric showed the same pattern. Guidance was effective in all six conditions, with the estimated marginal mean of $R$ significantly greater than zero in each condition (all $p<0.001$; Fig.~\ref{fig:emm_experiment_b}). Group size had a strong effect ($F_{1,18}=26.7$, $p<0.001$), with $R$ being higher at $N_r=5$ than at $N_r=8$. This decrease with increasing group size was significant within every configuration (fixed formation, $p=0.005$; independent $N_v=2$, $p=0.024$; independent $N_v=3$, $p=0.004$). In contrast, agent configuration had no significant effect ($F_{2,18}=1.55$, $p=0.24$), and there was no interaction between group-size and configuration ($F_{2,18}=0.20$, $p=0.82$). Although the independent configurations (Cluster-assignment mode) were intended to improve guidance by tracking subgroups, they did not differ significantly from the fixed-formation configuration (Global mode), which gave numerically the highest $R$ at each group size.

The positional histograms (Fig.~\ref{fig:hist_experiment_b}) reflect these results. At $N_r=5$ (Fig.~\ref{fig:hist_r5}), the centroid distributions for all three configurations were visibly shifted toward the target direction, with only minor differences among the modes (Bhattacharyya distances $0.124$, $0.076$, and $0.104$ for the fixed configuration and the $N_v=2$ and $N_v=3$ independent configurations, respectively). At $N_r=8$ (Fig.~\ref{fig:hist_r8}), the distributions were markedly more concentrated near the center across all configurations, and the Bhattacharyya distances were correspondingly lower ($0.076$, $0.047$, and $0.057$, respectively). Thus, although guidance remained statistically significant at $N_r=8$, its strength was substantially reduced, indicating reduced guidance efficacy at the larger group size.

\subsubsection{Temporal structure of the guidance response}
\label{sec:temporal_structure}

\begin{figure}[tbp]
  \centering

  \begin{subfigure}{\textwidth}
    \centering
    \begin{minipage}{\textwidth}
      \centering
      \makebox[0.23\textwidth][c]{\scriptsize Trial 1}\hfill
      \makebox[0.23\textwidth][c]{\scriptsize Trial 2}\hfill
      \makebox[0.23\textwidth][c]{\scriptsize Trial 3}\hfill
      \makebox[0.23\textwidth][c]{\scriptsize Trial 4}\par
      \vspace{1.5pt} 
      \includegraphics[width=0.23\textwidth]{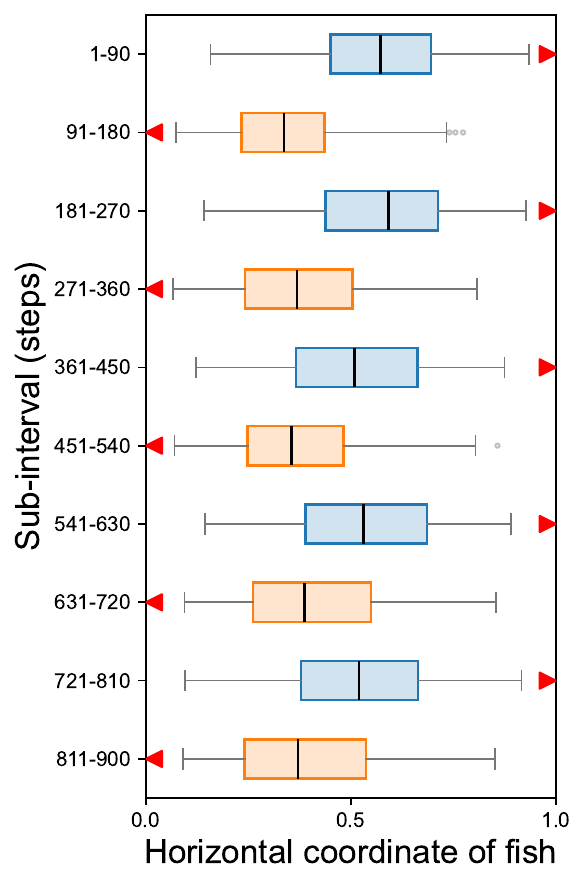}\hfill
      \includegraphics[width=0.23\textwidth]{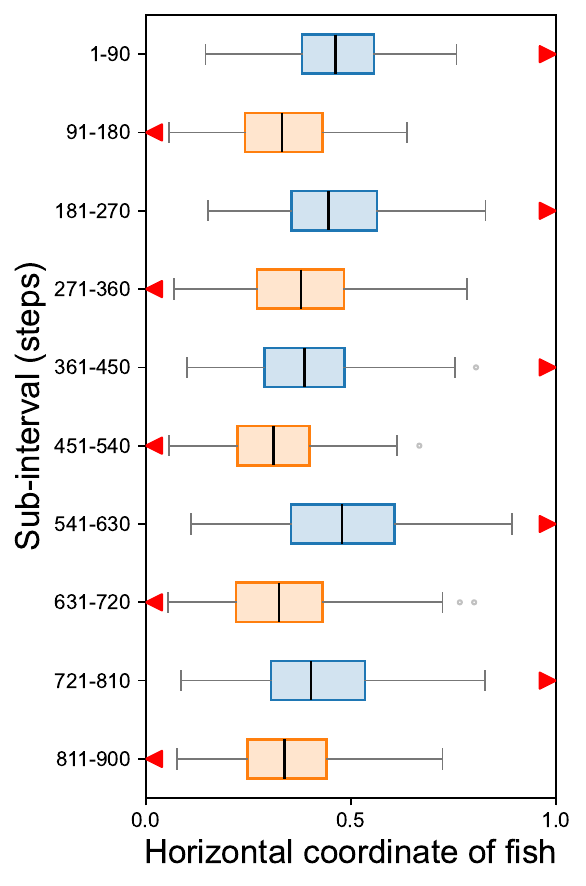}\hfill
      \includegraphics[width=0.23\textwidth]{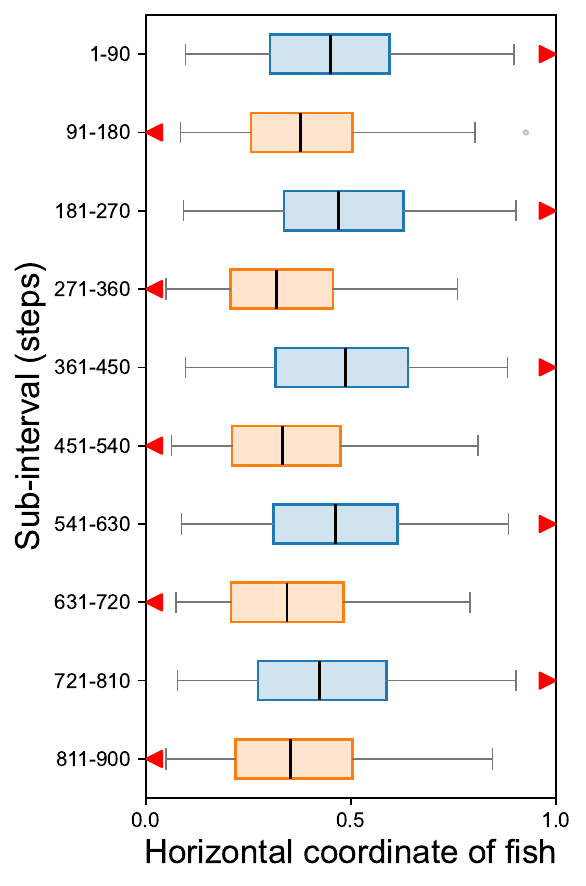}\hfill
      \includegraphics[width=0.23\textwidth]{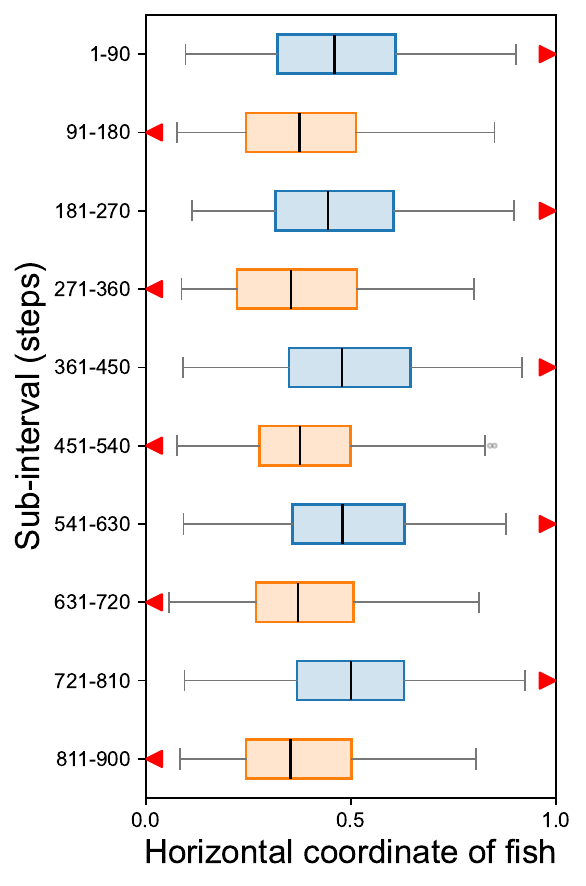}
    \end{minipage}
    \caption{Fixed formation}
    \label{fig:boxp_r5_fixed}
  \end{subfigure}

  \vspace{3ex}

  \begin{subfigure}{\textwidth}
    \centering
    \begin{minipage}{\textwidth}
      \centering
      \makebox[0.23\textwidth][c]{\scriptsize Trial 1}\hfill
      \makebox[0.23\textwidth][c]{\scriptsize Trial 2}\hfill
      \makebox[0.23\textwidth][c]{\scriptsize Trial 3}\hfill
      \makebox[0.23\textwidth][c]{\scriptsize Trial 4}\par
      \vspace{1.5pt}
      \includegraphics[width=0.23\textwidth]{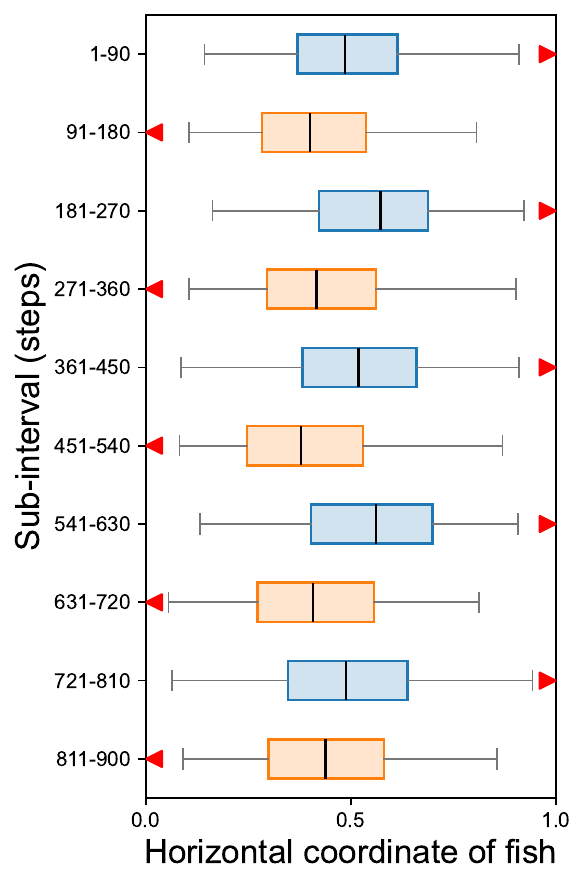}\hfill
      \includegraphics[width=0.23\textwidth]{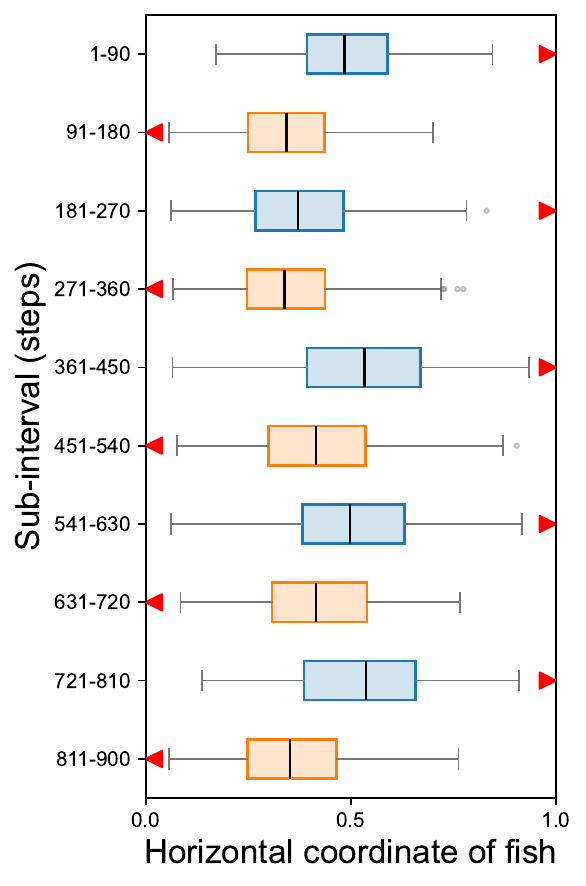}\hfill
      \includegraphics[width=0.23\textwidth]{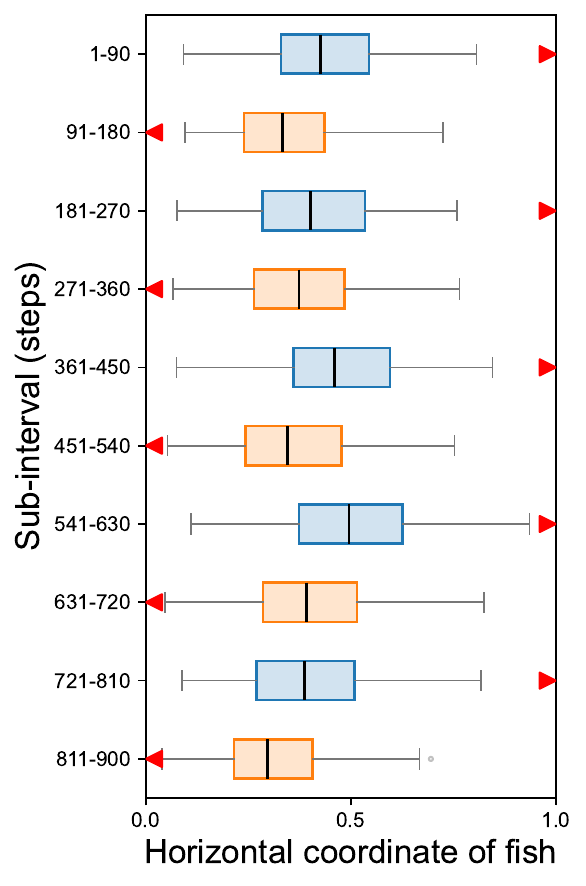}\hfill
      \includegraphics[width=0.23\textwidth]{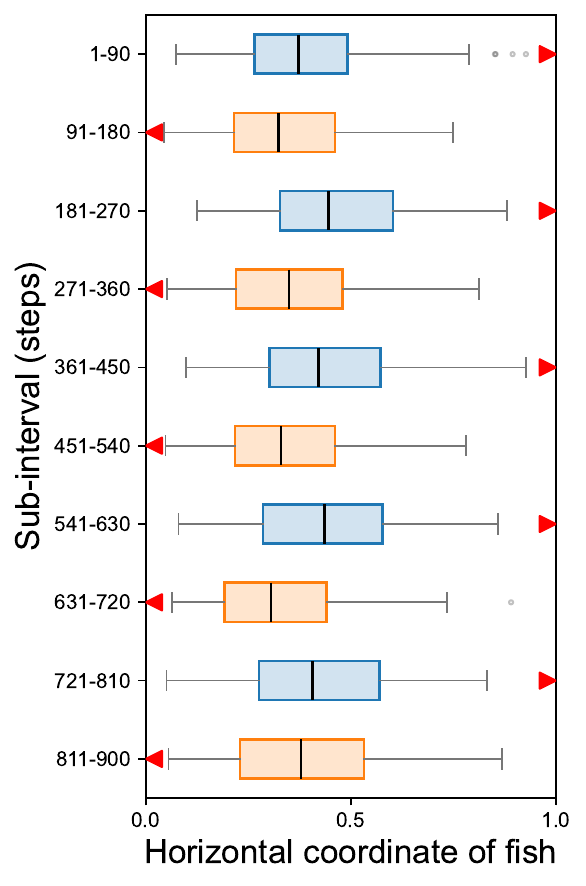}
    \end{minipage}
    \caption{Independent agents ($N_v = 2$)}
    \label{fig:boxp_r5_nv2}
  \end{subfigure}

  \caption{Distribution of individuals' x-coordinates in each sub-interval ($N_r=5$). Each row displays four trials for (a) the fixed formation and (b) the independent agent ($N_v=2$) configuration. Horizontal and vertical axes denote horizontal coordinates and elapsed sub-intervals, respectively. Target direction is indicated by red triangles and box colors (orange: leftward, blue: rightward).}  
  \label{fig:box_comparison_r5}
\end{figure}

To illustrate the guidance protocol and to inspect the temporal structure of the school's response, Fig.~\ref{fig:box_comparison_r5} shows examples of the distribution of the individuals' horizontal positions within each of the ten 90-step sub-intervals (approximately 1.8 minutes), for the four trials of the fixed-formation and the independent-agent ($N_v=2$) configurations.

The figure shows two consistent patterns.
First, the school's position tracked the alternating target direction across sub-intervals, confirming that the repeated reversals elicited a repeated behavioral response.
This tracking persisted throughout the trial, with no obvious decline visible, providing no clear indication of habituation to the virtual agents or fatigue.
Second, the response was asymmetric: the positional shift was consistently more pronounced during leftward guidance than during rightward guidance. This left--right asymmetry was observed in every condition, including those of Experiment~A, and was accounted for in the statistical models by the fixed effect of target direction (Section~\ref{sec:statistical_analysis}). Its possible causes are discussed in Section~\ref{sec:limitation_asymmetry}.

Notably, the two configurations (Figs.~\ref{fig:boxp_r5_fixed} and \ref{fig:boxp_r5_nv2}) produced similar sub-interval responses despite differing in the number and coordination of the agents, consistent with the absence of a significant configuration effect on $R$. 
This raises the question of how the agents themselves behaved while producing these responses, which we address in Section~\ref{sec:agent_behavior} by analyzing the motion of the virtual agents directly.

\subsection{Agent behavior}
\label{sec:agent_behavior}

To understand how the trained agents produced the responses in the physical experiments, we analyzed the virtual agents directly.  

\subsubsection{Distance-dependent behavior}
\label{sec:agent_direction}

In each step of the fixed-formation configuration, the displacement of the agent from control step $t$ to $t+1$ was projected onto two directions that reflect the composite reward (Eq.~\ref{eq:r_beta}): the \emph{toward-target} direction (the horizontal direction toward the current target end) and the \emph{toward-school} direction (the unit vector from the agent to the school centroid).
We report the probabilities that each projection is positive: $P(\text{toward target})$ and $P(\text{toward school})$. Note that the two directions are not mutually exclusive, because the school centroid may itself lie toward the target.

Figure~\ref{fig:agent_a} shows these probabilities as a function of the distance from agent to school centroid, which is denoted by $d$, for the fixed-formation agent at the three group sizes ($N_r = 3, 5, 8$; white background, large image).
The distances were divided into bins of width 0.05, and the probabilities were calculated by pooling the valid steps across trials within each group size.
For $N_r = 3$, the two white--large recording sets (background color and fish-image size experiments) were pooled. 
For this single-agent configuration the agent's reference point coincides with the global school centroid; therefore, $d$ is unambiguous. 
Across all three group sizes the behavior varied with distance in the same manner. When the agent was close to the school, its motion was predominantly directed toward the target, whereas at larger distances this probability fell toward zero and the motion was instead directed toward the school. 

In addition, among the steps when the agent was far from the school (here, $d > \theta$ was used), the fraction of steps in which the agent's horizontal position lay between the school centroid and the target end was computed in each trial. 
This fraction averaged more than $95\%$ in every group size ($95.4\%$, $98.0\%$, and $98.0\%$ for $N_r = 3, 5, 8$).
These results suggest that the agent acquired a distance-dependent policy through reinforcement learning: it moves toward the target while remaining near the school, and re-approaches the school once it has drawn too far ahead. This transition occurred at a comparable distance in all group sizes, close to the interaction range of simulated fish used during training ($\theta = 0.3$; Supplementary Table~\ref{tab:supp_dynamics_params}).

\subsubsection{Coordination among independent agents}
\label{sec:agent_coordination}

\begin{figure}[tb]
  \centering
  \begin{subfigure}[b]{0.56\textwidth}
    \centering
    \includegraphics[width=\textwidth]{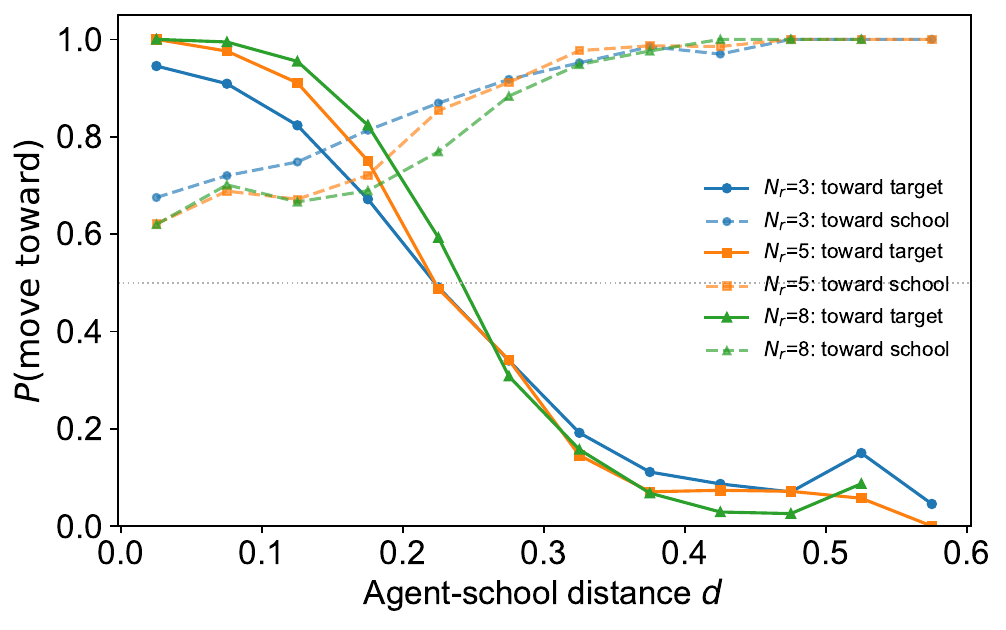}
    \caption{Distance-dependent behavior of fixed-formation agents}
    \label{fig:agent_a}
  \end{subfigure}
  \hfill
  \begin{subfigure}[b]{0.42\textwidth}
    \centering
    \includegraphics[width=\textwidth]{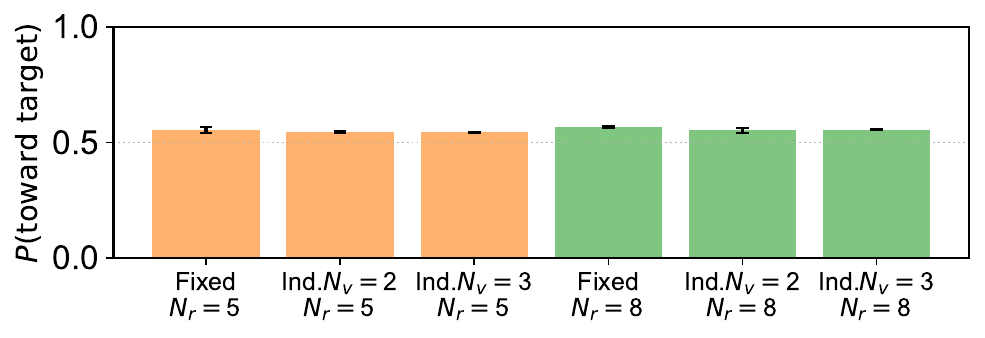}
    \caption{Target-directed motion (Exp.~B)}
    \label{fig:agent_b}

    \vspace{2ex}

    \includegraphics[width=\textwidth]{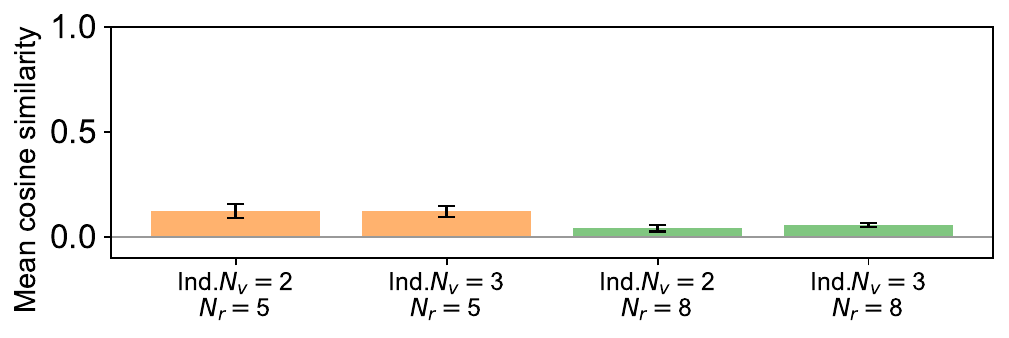}
    \caption{Inter-agent alignment (Exp.~B)}
    \label{fig:agent_c}
  \end{subfigure}
  \caption{Behavior of the virtual agents in the physical experiments.
    (a)~Probability that the fixed-formation agent moves toward the target end (solid) and toward the school centroid (dashed) as a function of the agent--school distance $d$, for group sizes $N_r = 3, 5, 8$; the $N_r = 3$ curve pools the two white--large recording sets (the white and large conditions of Experiment~A). 
    (b)~Probability that an agent's displacement has a positive target-directed component, pooled over agents and steps, for the six Experiment~B configurations.
    (c)~Mean pairwise cosine similarity of the agents' displacement vectors for the four independent configurations.
    In (b) and (c), mean values across four trials are shown, with SD indicated by error bars.}
  \label{fig:agent_behavior}
\end{figure}

We next compared the agents across the six Experiment~B configurations.
As the independent configurations were designed to track clusters of fish, the use of the global school centroid as a reference point may be inappropriate for comparing agent behavior; we therefore considered only the probability that each agent moved toward the target, i.e., $P(\text{toward target})$. This probability was essentially the same across all configurations (${\approx}\,0.55$), as shown in Fig.~\ref{fig:agent_b}, which suggests that each agent, whether acting as a single fixed formation or as one of several independent agents, pursued the target to the same degree.

However, the configurations differed in how coherently the agents moved. For the independent configurations we measured the mean pairwise cosine similarity of the agents' displacement vectors. The alignment was low in every independent configuration (Fig.~\ref{fig:agent_c}): the mean cosine similarity was about $0.12$ at $N_r = 5$ and only about $0.05$ at $N_r = 8$.

This small alignment is attributable to the target drive shared by all agents (Fig.~\ref{fig:agent_b}), while the agents likely moved with different timing.
Beyond this common drive, the agents were essentially uncorrelated, consistent with their independent design: each agent re-approached the centroid of its own assigned cluster rather than a common point.

\section{Discussion}

\subsection{Effectiveness of the multi-objective reward design}

A key contribution of our reinforcement learning framework is a composite reward for fish-school guidance that introduces a form of functional biomimicry at the level of the control objective.
The cohesion term encourages the virtual agent to remain close to the school, thereby preserving the interaction required to influence its motion, whereas the directional term encourages the agent to progress toward the prescribed target.

The simulation results support the importance of this balance.
Policies trained with $\beta=0.3$ outperformed the previously used single-reward baseline $r_\mathrm{base}$ at $T=10^6$ steps under all tested stochastic conditions except for the deterministic case $p=0$, whereas other reward weights produced performance comparable to the baseline (Fig.~\ref{fig:simulation_result}).
This result indicates that the relative weighting of cohesion and directional progress affects guidance performance.

The selected policy also produced significant target-directed guidance after zero-shot deployment in the physical experiments under suitable visual conditions, suggesting that the interaction strategy acquired under the stochastic simulation model remained effective with live fish.
The behavior of the agents in the physical trials (Section~\ref{sec:agent_direction}) further suggests that the learned behavior was consistent with both objectives: the agent moved toward the target while remaining close to the school, but re-approached the school after advancing too far ahead.
This distance-dependent behavior was common to all tested group sizes.
Together, these findings support the use of a functionally structured reward design for closed-loop interaction with biological collectives.

\subsection{Optimization of visual stimuli for collective guidance}

The results of Experiment~A showed that the visual presentation conditions strongly affected the guidance efficacy of the virtual agents.
Among the tested conditions, the white background and the large fish-image size produced the highest guidance efficacy for rummy-nose tetras.
This finding emphasizes that the effectiveness of closed-loop control depends not only on the learned policy but also on whether the resulting actions are presented as sufficiently salient visual stimuli.

Guidance was significant under the white and gray backgrounds but not under the black background, under which the school remained predominantly in the intermediate area rather than shifting toward the target area (Table~\ref{tab:experiment_a_results}, Fig.~\ref{fig:hist_bg}).
One possible explanation is that the white and gray backgrounds provided visual conditions under which the fish images were more readily detectable or distinguishable from their surroundings.
By contrast, the reduced guidance under the black background may reflect lower stimulus visibility, an effect of ambient brightness on fish behavior, or a combination of these factors.
We note that the luminance contrast between the images and each background was not quantified in this study; directly measuring this contrast and distinguishing its effect from that of ambient brightness remain important topics for future work.

Regarding fish-image size, the largest size, corresponding to approximately $1.5$ times the size of real individuals, produced the strongest guidance response among the three tested sizes, whereas no significant guidance was detected for the small images (Table~\ref{tab:experiment_a_results}, Fig.~\ref{fig:emm_experiment_a} right).
While this might suggest that larger-than-real stimuli are more effective, direct comparisons with the size of real individuals should be interpreted with caution, as the virtual agents were presented on a screen with a minimum separation of approximately 47~mm from the fish.
Within the range tested, however, guidance increased with image size (Fig.~\ref{fig:hist_size}), suggesting that larger stimuli may provide a more salient cue and thereby promote collective directional changes.

\subsection{Group-size-dependent limitations of visual guidance}

Guidance efficacy decreased significantly as group size increased from $N_r=5$ to $N_r=8$ across all tested agent configurations (Table~\ref{tab:experiment_b_results}, Fig.~\ref{fig:hist_experiment_b}, and Supplementary Table~\ref{tab:supp_pairwise}).
Although significant target-directed guidance was retained for schools of eight fish, the positional distributions became more centralized and less clearly shifted toward the target than those observed for schools of five fish.
These results indicate that maintaining artificial influence becomes increasingly difficult as the number of individuals in a school increases.

Because the observed agent behavior was broadly comparable across group sizes (Fig.~\ref{fig:agent_behavior}), this reduction may reflect increasing competition between the influence of the virtual agents and interactions among real individuals.
In larger schools, social information from neighboring fish may reduce the relative influence of external visual stimuli and limit the propagation of agent-induced motion through the group.
Hydrodynamic cues detected through the lateral line may also become more influential as more fish occupy the same space~\cite{liVortexPhaseMatching2020,koRoleHydrodynamicsCollective2023}.
Because these mechanisms were not directly measured, future work should examine how visual, social, and hydrodynamic influences, as well as the spatial cohesion of the school, change with group size and density.

\subsection{Challenges in multi-agent control and coordination}

The independent multi-agent configurations ($N_v = 2, 3$), although intended to improve guidance by tracking subgroups, did not outperform the fixed-formation configuration (Global mode) (Table~\ref{tab:experiment_b_results}).
This lack of improvement may be explained by several factors.
First, the fixed-formation configuration utilized four fish images controlled as a single unit, whereas the independent configurations used only two or three fish images (Section~\ref{sec:experimental_design}).
This larger number of visual stimuli in the fixed-formation configuration likely increased overall salience, providing a more robust and readily recognizable directional signal.

Beyond stimulus strength, the lack of coordination among the agents may also have contributed.
Our analysis of agent behavior indicates that the shortfall was not in the individual policies: each agent pursued the target to a degree comparable to that of the fixed-formation agent (Fig.~\ref{fig:agent_b}).
What the independent configurations lacked was coherence among the agents, whose displacements showed little correlation with one another (Fig.~\ref{fig:agent_c}).
Because the policy was trained in a single-agent setting and each agent acted only on its own position and assigned reference point (Section~\ref{sec:state_action}), the agents had no information about one another and hence no means of coordinating their actions.
The resulting stimulus may therefore have appeared to the school as several loosely related moving objects rather than as a single unified group, thereby weakening the directional signal conveyed to the fish.

These findings suggest that simply increasing the number of independently controlled agents is insufficient to improve guidance.
Guiding fragmented or heterogeneous groups may instead require cooperative multi-agent reinforcement learning (MARL)~\cite{MAPPO2021}, in which agents coordinate their actions to provide a coherent guidance signal.
Future comparisons should also control for the total number of displayed fish images to separate the effects of stimulus salience from those of agent coordination.

\subsection{Limitations}
\label{sec:limitation_asymmetry}

A slight positional bias toward the left side of the tank was observed across the physical experiments, as described in Section~\ref{sec:temporal_structure}.
Because $R$ is defined relative to the tank center (Eq.~\ref{eq:r_base}), such an offset inflates $R$ for leftward guidance and deflates it for rightward guidance.
The balanced alternation of target direction causes these opposing effects to largely cancel, and target direction was additionally included as a fixed effect in the statistical analysis (Section~\ref{sec:statistical_analysis}).
The reported estimates therefore primarily reflect the guidance effect rather than the static positional bias.
The origin of this bias was not identified. It may stem from subtle environmental factors such as inhomogeneous lighting, feeding habituation, or asymmetries in the housing tanks, and it should be mitigated in future studies.

The experimental design also limits the scope of our conclusions.
Our comparison of group sizes was restricted to $N_r=5$ and $N_r=8$, and each condition was assessed with only four trials.
Because each trial drew fish from a small holding population (five for Experiment~A and twelve for Experiment~B), the same individuals recurred across trials. 
The trials within a condition are therefore not fully independent replicates, and because this non-independence could not be modeled (Section~\ref{sec:statistical_analysis}), the $p$-values, especially those close to the significance threshold, should be interpreted with this in mind.

The evaluation also depends on the accuracy of the tracking system.
We removed only static false detections from the tracking data (Section~\ref{sec:eval_metrics}), and non-static false positives, if present, were not identified.
The detection rate was slightly lower when the fish were aggregated near the tank ends, where individuals overlap more frequently.
Because such aggregation coincided with successful guidance, this may have led to an underestimation of guidance efficacy, although the magnitude of this effect was not quantified.

Finally, the virtual agents were presented solely as two-dimensional visual stimuli and provided no hydrodynamic feedback, which may have limited their ability to influence the school, particularly in larger groups.

\section{Conclusion}
\label{sec:conclusion}

In this study, we proposed and evaluated a deep reinforcement learning framework for the real-time, closed-loop guidance of fish schools.
By employing Proximal Policy Optimization (PPO) and a composite reward function that balances directional guidance with cohesion, we developed an autonomous controller capable of guiding biological collectives.
Our methodology demonstrates a bridge between simulation-based training and real-world application, achieving effective zero-shot transfer to physical experiments with rummy-nose tetras.
An analysis of the agents' motion in these experiments showed that the transferred policy retained the intended balance between the two reward terms, with the agents moving toward the target while remaining close to the school and re-approaching the school after advancing too far ahead.

Our findings identify important factors governing the efficacy of artificial social influence.
We found that the salience of visual stimuli, specifically background color and stimulus size, played a major role in maximizing the responsiveness of the fish school. 
Furthermore, our evaluation across group sizes and agent configurations highlights a key limitation: while virtual agents can effectively guide smaller groups, their influence diminishes in larger groups, where intrinsic social interactions and sensory competition may increasingly outweigh the external visual stimuli.
Increasing the number of independently controlled agents did not improve guidance. Although each agent pursued the target as strongly as the fixed-formation agent did, the agents moved almost independently of one another, indicating that a lack of coordination may limit the effectiveness of such configurations, even when individual agents are well optimized.

This work establishes a foundation for the automated guidance of biological groups. Future research will focus on implementing cooperative multi-agent reinforcement learning (MARL) to facilitate the coordination of agent actions in response to diverse group dynamics, including fragmented subgroups. Such coordination will be necessary to provide a coherent and robust social signal and, alongside the integration of multimodal stimuli such as hydrodynamic feedback, to maintain influence in dense and complex biological environments.

\section*{Acknowledgments}
The authors are grateful to Yusuke Nishii for developing the foundational framework~\cite{Nishii18-26} upon which this study was built.

\subsection*{Funding}
This study was supported by the JSPS KAKENHI Grant Number JP21H05302.

\subsection*{Conflicts of interest}
The authors declare no personal or financial competing interests.

\section*{Data availability}
The data and analysis scripts are available at \url{https://doi.org/10.6084/m9.figshare.32749785} (reserved; will be made publicly available upon publication)~\cite{DataAndScripts2026}.

\section*{Ethical statement}
All animal experiments were conducted with the approval of the Graduate School of Information Science, University of Hyogo, Japan (Approval No. UHIS-EC-2024-004).

\printbibliography

\clearpage
\setcounter{page}{1}
\renewcommand{\thepage}{S\arabic{page}}

\setcounter{table}{0}
\renewcommand{\thetable}{S\arabic{table}}

\begin{center}
{\Large\bfseries Supplementary Tables\par}

\vspace{2em}

Takato Shibayama and Hiroaki Kawashima\textsuperscript{*}\\
Graduate School of Information Science, University of Hyogo, Kobe, Japan

\vspace{0.4em}

{\small \textsuperscript{*}Corresponding author: kawashima@gsis.u-hyogo.ac.jp}
\end{center}

\vspace{2em}
\noindent\textbf{Description:}\\
Additional implementation details, trial-level data, and statistical results supporting the analysis of closed-loop fish-school guidance experiments.

\vspace{1em}
\noindent\textbf{Supplementary material for:} A Deep Reinforcement Learning Framework for Closed-loop Guidance of Fish Schools via Virtual Agents

\vspace{2em}
\begin{table}[htbp]
\centering
\caption{Parameters used for the virtual-agent and simulated-school dynamics. The inverse time constants are the corresponding forward-Euler update coefficients multiplied by the update frequency.}
\label{tab:supp_dynamics_params}
\begin{tabular}{ll}
\toprule
Parameter & Value \\
\midrule
\multicolumn{2}{l}{\textit{Simulation: virtual agent}} \\
\quad Displacement $D$ & 0.2 \\
\quad Inverse time constant $1/\tau^{(v)}$ & \qty{3}{\per\s} \\
\addlinespace
\multicolumn{2}{l}{\textit{Simulation: simulated school}} \\
\quad Maximum phase duration $\Delta t_{\max}$ & \qty{3}{\s} \\
\quad Inverse time constant $1/\tau^{(r)}$ & \qty{1}{\per\s} \\
\quad Interaction range $\theta$ & 0.3 \\
\quad Maximum horizontal random displacement $\delta x_{\max}$ & 0.2 \\
\quad Maximum vertical random displacement $\delta y_{\max}$ & 0.2 \\
\addlinespace
\multicolumn{2}{l}{\textit{Physical experiments: virtual agent}} \\
\quad Displacement $D$ & 0.3 \\
\quad Inverse time constant $1/\tau^{(v)}$ & \qty{2.25}{\per\s} \\
\bottomrule
\end{tabular}
\end{table}

\vspace{1em}
\begin{table}[htbp]
\centering
\caption{PPO hyperparameters used for policy training.}
\label{tab:supp_ppo_params}
\begin{tabular}{p{0.50\linewidth} p{0.40\linewidth}}
\toprule
Parameter & Value \\
\midrule
Network architecture & 96-unit shared hidden layer (ReLU);\newline separate policy/value heads \\
Optimizer & Adam \\
Learning rate & $5 \times 10^{-4}$ \\
Discount factor $\gamma$ & 0.15 \\
Policy update interval & 5 steps \\
Minibatch size for gradient descent & 10 \\
Number of gradient-descent epochs per update & 1 \\
\bottomrule
\end{tabular}
\end{table}

\newpage
\begin{longtable}{llrrrrr}
\caption{Per-trial values (four trials per condition). Target and opposite occupancies are the percentages of detected individuals located in the target and opposite areas (pooled over steps). $R$ is the per-trial guidance metric (temporal average over the trial). Det.\ rate is the mean per-step detection rate, i.e.\ the number of localized fish divided by the group size $N_r$, averaged over all steps; it reflects missed (undetected) individuals after removal of static false positives (Section~\ref{sec:eval_metrics}). Valid steps is the number of steps with at least one detected individual, i.e.\ the steps contributing to $R$ and the occupancy ratios. The \textit{Mean $\pm$ SD} row in each block gives the condition mean, matching the corresponding main-text summary tables (Tables~\ref{tab:experiment_a_results} and ~\ref{tab:experiment_b_results}).}\label{tab:supp_trials}\\
\toprule
Condition & Trial & Target (\%) & Opposite (\%) & $R$ & Det.\ rate & Valid steps \\
\midrule
\endfirsthead
\multicolumn{7}{l}{\tablename\ \thetable\ (continued)}\\
\toprule
Condition & Trial & Target (\%) & Opposite (\%) & $R$ & Det.\ rate & Valid steps \\
\midrule
\endhead
\bottomrule
\endfoot
\bottomrule
\endlastfoot
\multicolumn{7}{l}{\textit{Experiment A}}\\
\midrule
Background: white & 1 & 33.5 & 9.9 & 0.17 & 0.932 & 898 \\
 & 2 & 27.2 & 8.9 & 0.14 & 0.933 & 899 \\
 & 3 & 27.1 & 5.3 & 0.18 & 0.860 & 900 \\
 & 4 & 30.0 & 5.4 & 0.18 & 0.939 & 900 \\
\cmidrule(lr){3-7}
\multicolumn{2}{r}{\textit{Mean $\pm$ SD}} & $29.4 \pm 3.0$ & $7.4 \pm 2.4$ & $0.17 \pm 0.02$ & 0.916 & 899 \\
\midrule
Background: gray & 1 & 26.1 & 8.0 & 0.14 & 0.914 & 899 \\
 & 2 & 24.9 & 5.6 & 0.15 & 0.869 & 900 \\
 & 3 & 28.6 & 9.4 & 0.15 & 0.896 & 900 \\
 & 4 & 21.9 & 5.4 & 0.13 & 0.947 & 900 \\
\cmidrule(lr){3-7}
\multicolumn{2}{r}{\textit{Mean $\pm$ SD}} & $25.4 \pm 2.8$ & $7.1 \pm 1.9$ & $0.14 \pm 0.01$ & 0.906 & 900 \\
\midrule
Background: black & 1 & 12.9 & 7.8 & 0.04 & 0.910 & 898 \\
 & 2 & 23.5 & 17.2 & 0.04 & 0.950 & 900 \\
 & 3 & 6.4 & 6.5 & $-$0.01 & 0.934 & 900 \\
 & 4 & 7.2 & 4.8 & 0.01 & 0.960 & 900 \\
\cmidrule(lr){3-7}
\multicolumn{2}{r}{\textit{Mean $\pm$ SD}} & $12.5 \pm 7.9$ & $9.1 \pm 5.6$ & $0.02 \pm 0.02$ & 0.939 & 900 \\
\midrule
Size: small & 1 & 24.2 & 14.9 & 0.07 & 0.960 & 900 \\
 & 2 & 18.7 & 13.6 & 0.03 & 0.996 & 900 \\
 & 3 & 21.2 & 20.5 & 0.00 & 0.940 & 900 \\
 & 4 & 16.5 & 22.5 & $-$0.02 & 0.906 & 899 \\
\cmidrule(lr){3-7}
\multicolumn{2}{r}{\textit{Mean $\pm$ SD}} & $20.2 \pm 3.3$ & $17.9 \pm 4.3$ & $0.02 \pm 0.04$ & 0.950 & 900 \\
\midrule
Size: medium & 1 & 30.8 & 13.8 & 0.13 & 0.948 & 900 \\
 & 2 & 22.7 & 10.0 & 0.10 & 0.966 & 900 \\
 & 3 & 11.4 & 11.9 & $-$0.00 & 0.998 & 900 \\
 & 4 & 31.0 & 11.1 & 0.15 & 0.938 & 898 \\
\cmidrule(lr){3-7}
\multicolumn{2}{r}{\textit{Mean $\pm$ SD}} & $24.0 \pm 9.2$ & $11.7 \pm 1.6$ & $0.10 \pm 0.07$ & 0.962 & 900 \\
\midrule
Size: large & 1 & 36.3 & 6.2 & 0.23 & 0.901 & 899 \\
 & 2 & 36.4 & 10.2 & 0.19 & 0.932 & 900 \\
 & 3 & 14.3 & 9.8 & 0.03 & 0.989 & 900 \\
 & 4 & 38.1 & 7.5 & 0.24 & 0.924 & 893 \\
\cmidrule(lr){3-7}
\multicolumn{2}{r}{\textit{Mean $\pm$ SD}} & $31.3 \pm 11.3$ & $8.4 \pm 1.9$ & $0.17 \pm 0.10$ & 0.936 & 898 \\
\newpage
\multicolumn{7}{l}{\textit{Experiment B}}\\
\midrule
$N_r$=5, Fixed & 1 & 29.5 & 7.6 & 0.16 & 0.967 & 900 \\
 & 2 & 22.6 & 9.0 & 0.10 & 0.951 & 900 \\
 & 3 & 26.3 & 13.3 & 0.10 & 0.953 & 900 \\
 & 4 & 24.3 & 10.9 & 0.10 & 0.968 & 900 \\
\cmidrule(lr){3-7}
\multicolumn{2}{r}{\textit{Mean $\pm$ SD}} & $25.7 \pm 3.0$ & $10.2 \pm 2.4$ & $0.12 \pm 0.03$ & 0.960 & 900 \\
\midrule
$N_r$=5, Ind. $N_v$=2 & 1 & 24.1 & 9.8 & 0.10 & 0.977 & 900 \\
 & 2 & 22.2 & 8.3 & 0.11 & 0.940 & 900 \\
 & 3 & 21.9 & 11.1 & 0.09 & 0.958 & 900 \\
 & 4 & 25.0 & 15.1 & 0.08 & 0.948 & 900 \\
\cmidrule(lr){3-7}
\multicolumn{2}{r}{\textit{Mean $\pm$ SD}} & $23.3 \pm 1.5$ & $11.1 \pm 2.9$ & $0.09 \pm 0.01$ & 0.956 & 900 \\
\midrule
$N_r$=5, Ind. $N_v$=3 & 1 & 29.5 & 9.8 & 0.14 & 0.973 & 900 \\
 & 2 & 21.8 & 9.1 & 0.10 & 0.954 & 900 \\
 & 3 & 25.9 & 13.6 & 0.09 & 0.951 & 900 \\
 & 4 & 25.9 & 11.0 & 0.11 & 0.941 & 900 \\
\cmidrule(lr){3-7}
\multicolumn{2}{r}{\textit{Mean $\pm$ SD}} & $25.8 \pm 3.1$ & $10.9 \pm 2.0$ & $0.11 \pm 0.02$ & 0.955 & 900 \\
\midrule
$N_r$=8, Fixed & 1 & 23.0 & 7.9 & 0.11 & 0.904 & 900 \\
 & 2 & 19.1 & 9.0 & 0.07 & 0.944 & 900 \\
 & 3 & 22.1 & 14.1 & 0.06 & 0.952 & 900 \\
 & 4 & 19.1 & 12.1 & 0.04 & 0.955 & 900 \\
\cmidrule(lr){3-7}
\multicolumn{2}{r}{\textit{Mean $\pm$ SD}} & $20.8 \pm 2.0$ & $10.8 \pm 2.8$ & $0.07 \pm 0.03$ & 0.939 & 900 \\
\midrule
$N_r$=8, Ind. $N_v$=2 & 1 & 18.6 & 10.4 & 0.06 & 0.926 & 900 \\
 & 2 & 23.0 & 11.9 & 0.08 & 0.953 & 900 \\
 & 3 & 19.0 & 13.5 & 0.04 & 0.968 & 900 \\
 & 4 & 21.2 & 14.2 & 0.05 & 0.958 & 900 \\
\cmidrule(lr){3-7}
\multicolumn{2}{r}{\textit{Mean $\pm$ SD}} & $20.5 \pm 2.1$ & $12.5 \pm 1.7$ & $0.06 \pm 0.02$ & 0.951 & 900 \\
\midrule
$N_r$=8, Ind. $N_v$=3 & 1 & 19.3 & 9.5 & 0.07 & 0.912 & 900 \\
 & 2 & 22.9 & 12.6 & 0.08 & 0.944 & 900 \\
 & 3 & 19.5 & 12.1 & 0.05 & 0.951 & 900 \\
 & 4 & 22.1 & 17.2 & 0.04 & 0.937 & 900 \\
\cmidrule(lr){3-7}
\multicolumn{2}{r}{\textit{Mean $\pm$ SD}} & $20.9 \pm 1.8$ & $12.9 \pm 3.2$ & $0.06 \pm 0.02$ & 0.936 & 900 \\
\end{longtable}

\begin{table}[ht]
\centering
\caption{Fixed-effect Type~III $F$-tests (Kenward--Roger denominator d.f.).} 
\label{tab:supp_ftests}
\begin{tabular}{llrrrl}
\toprule
Model & Effect & NumDF & DenDF & $F$ & $p$ \\
\midrule
\multirow{2}{*}{\shortstack[l]{\textit{Exp.~A}\\Background color}} & level & 2 & 9.00 & 47.02 & $<0.001$ \\
 & target\_dir & 1 & 11.00 & 49.20 & $<0.001$ \\
\midrule
\multirow{2}{*}{\shortstack[l]{\textit{Exp.~A}\\Fish-image size}} & level & 2 & 9.00 & 4.45 & 0.045 \\
 & target\_dir & 1 & 11.00 & 24.81 & $<0.001$ \\
\midrule
\multirow{4}{*}{\shortstack[l]{\textit{Exp.~B}\\Group size $\times$ configuration}} & group\_size & 1 & 18.00 & 26.67 & $<0.001$ \\
 & config & 2 & 18.00 & 1.55 & 0.240 \\
 & target\_dir & 1 & 23.00 & 75.91 & $<0.001$ \\
 & group\_size:config & 2 & 18.00 & 0.20 & 0.817 \\
\bottomrule
\end{tabular}
\end{table}

\begin{table}[ht]
\centering
\caption{Estimated marginal means of $R$ (averaged over target direction), with 95\% Kenward--Roger CIs and the test against zero.} 
\label{tab:supp_emm}
\begin{tabular}{llrrll}
\toprule
Experiment & Condition & $R$ & SE & 95\% CI & $p$ \\
\midrule
\multirow{3}{*}{\shortstack[l]{\textit{Exp.~A}\\Background color}} & white & 0.17 & 0.01 & [0.143, 0.195] & $<0.001$ \\
 & gray & 0.14 & 0.01 & [0.115, 0.168] & $<0.001$ \\
 & black & 0.02 & 0.01 & [$-$0.006, 0.046] & 0.114 \\
\midrule
\multirow{3}{*}{\shortstack[l]{\textit{Exp.~A}\\Fish-image size}} & small & 0.02 & 0.04 & [$-$0.062, 0.102] & 0.587 \\
 & medium & 0.10 & 0.04 & [0.014, 0.178] & 0.027 \\
 & large & 0.17 & 0.04 & [0.091, 0.255] & $<0.001$ \\
\midrule
\multirow{6}{*}{\textit{Exp.~B}} & $N_r$=5, Fixed & 0.12 & 0.01 & [0.095, 0.14] & $<0.001$ \\
 & $N_r$=5, Ind.$N_v$=2 & 0.09 & 0.01 & [0.07, 0.116] & $<0.001$ \\
 & $N_r$=5, Ind.$N_v$=3 & 0.11 & 0.01 & [0.087, 0.132] & $<0.001$ \\
 & $N_r$=8, Fixed & 0.07 & 0.01 & [0.046, 0.091] & $<0.001$ \\
 & $N_r$=8, Ind.$N_v$=2 & 0.06 & 0.01 & [0.033, 0.078] & $<0.001$ \\
 & $N_r$=8, Ind.$N_v$=3 & 0.06 & 0.01 & [0.037, 0.082] & $<0.001$ \\
\bottomrule
\end{tabular}
\end{table}

\begin{table}[ht]
\centering
\caption{Pairwise contrasts between conditions (Holm-adjusted).} 
\label{tab:supp_pairwise}
\begin{tabular}{lllrrrl}
\toprule
Set &  & Contrast & Estimate & SE & $t$ & $p$ \\
\midrule
\multirow{3}{*}{\shortstack[l]{\textit{Exp.~A}\\Background color}} &  & white - gray & 0.03 & 0.02 & 1.67 & 0.130 \\
 &  & white - black & 0.15 & 0.02 & 9.11 & $<0.001$ \\
 &  & gray - black & 0.12 & 0.02 & 7.44 & $<0.001$ \\
\midrule
\multirow{3}{*}{\shortstack[l]{\textit{Exp.~A}\\Fish-image size}} &  & small - medium & $-$0.07 & 0.05 & $-$1.47 & 0.327 \\
 &  & small - large & $-$0.15 & 0.05 & $-$2.98 & 0.046 \\
 &  & medium - large & $-$0.08 & 0.05 & $-$1.52 & 0.327 \\
\midrule
\multirow{3}{*}{\shortstack[l]{\textit{Exp.~B}\\(by configuration)}} & Fixed & $N_r$=5 - $N_r$=8 & 0.05 & 0.01 & 3.21 & 0.005 \\
\cmidrule(lr){2-7}
 & Ind.$N_v$=2 & $N_r$=5 - $N_r$=8 & 0.04 & 0.01 & 2.46 & 0.024 \\
\cmidrule(lr){2-7}
 & Ind.$N_v$=3 & $N_r$=5 - $N_r$=8 & 0.05 & 0.01 & 3.28 & 0.004 \\
\midrule
\multirow{6}{*}{\shortstack[l]{\textit{Exp.~B}\\(by group size)}} & \multirow{3}{*}{$N_r$=5} & Fixed - Ind.$N_v$=2 & 0.02 & 0.01 & 1.61 & 0.371 \\
 &  & Fixed - Ind.$N_v$=3 & 0.01 & 0.01 & 0.53 & 0.603 \\
 &  & Ind.$N_v$=2 - Ind.$N_v$=3 & $-$0.02 & 0.01 & $-$1.09 & 0.584 \\
\cmidrule(lr){2-7}
 & \multirow{3}{*}{$N_r$=8} & Fixed - Ind.$N_v$=2 & 0.01 & 0.01 & 0.87 & 1.000 \\
 &  & Fixed - Ind.$N_v$=3 & 0.01 & 0.01 & 0.60 & 1.000 \\
 &  & Ind.$N_v$=2 - Ind.$N_v$=3 & $-$0.00 & 0.01 & $-$0.27 & 1.000 \\
\bottomrule
\end{tabular}
\end{table}

\end{document}